# Deep Learning for Political Science[1]

Kakia Chatsiou (University of Essex) and Slava Jankin Mikhaylov (Hertie School)[2]

## Introduction

Political science, and social science in general, have traditionally been using computational methods to study areas such as voting behavior, policy making, international conflict, and international development. More recently, increasingly available quantities of data are being combined with improved algorithms and affordable computational resources to predict, learn, and discover new insights from data that is large in volume and variety. New developments in the areas of machine learning, deep learning, natural language processing (NLP), and, more generally, artificial intelligence (AI) are opening up new opportunities for testing theories and evaluating the impact of interventions and programs in a more dynamic and effective way. Applications using large volumes of structured and unstructured data are becoming common in government and industry, and increasingly also in social science research.

This chapter offers an introduction to such methods drawing examples from political science. Focusing on the areas where the strengths of the methods coincide with challenges in these fields, the chapter first presents an introduction to AI and its core technology – machine learning, with its rapidly developing subfield of deep learning. The discussion of deep neural networks is illustrated with the NLP tasks that are relevant to political science. The latest advances in deep learning methods for NLP are also reviewed, together with their potential for improving information extraction and pattern recognition from political science texts.

[2] Email address: jankin@hertie-school.org

We conclude by reflecting on issues of algorithmic bias – often overlooked in political science research. We also discuss the issues of fairness, accountability, and transparency in machine learning, which are being addressed at the academic and public policy levels.

## AI: Machine Learning and NLP

The European Commission (2019) defines AI as 'systems that display intelligent behaviour by analysing their environment and taking actions – with some degree of autonomy – to achieve specific goals'. As a scientific discipline, AI includes several techniques like machine learning (with deep learning and reinforcement learning as specific examples), machine reasoning, and robotics (European Commission, 2019). However, much of what is discussed as AI in the public sphere is machine learning, which is an 'algorithmic field that blends ideas from statistics, computer science and many other disciplines […] to design algorithms that process data, make predictions, and help make decisions' (Jordan, 2019).

Machine learning has a history of successful deployment in both industry and academia, going back several decades. Deep learning has more recently made great progress in such applications as speech and language understanding, computer vision, and event and behavior prediction (Goodfellow et al., 2016). These rapid technological advances and the promise of automation and human-intelligence augmentation (Jordan, 2019) reignited debates on AI's impact on jobs and markets (Brynjolfsson et al., 2018; Samothrakis, 2018; Schlogl and Sumner, 2018) and the need for AI governance (Aletras et al., 2016; Benjamins et al., 2005).

Machine learning (and deep learning as its subfield) is defined as the ‘field of study that gives computers the ability to learn without being explicitly programmed’ (Samuel, 1959). In this context, ‘learning’ can be viewed as the use of statistical techniques to enable computer systems to progressively improve their performance on a specific task using data without being explicitly programmed (Goldberg and Holland, 1988). To be able to learn how to perform a task and become better at it, a machine should:

- be provided with a set of example information (inputs) and the desired outputs. The goal is then to learn a general rule that can take us from the inputs to the outputs. This type of learning is called *Supervised Learning.* This works well even in cases when the input information is not available in full;
- be provided with an incomplete set of example information to learn from, where some of the target outputs are missing. This type of learning is called *Semi-supervised Learning*. When example information is available in one domain and we want to apply the knowledge to another domain with no available example information, this is called *Transfer Learning*;
- obtain training labels for a small number of instances while at the same time optimize which elements it needs to learn labels for. This is called *Active Learning*, and, in some cases, it can be implemented interactively in order to ask a human user for information on how best to label different elements;
- be asked to find structure in the input without having any labels provided in advance (as input). This type of learning is called *Unsupervised Learning* and can be used both for discovering hidden patterns in the data as well as learning features or parameters from the data;

be given information not about the structure of the data itself but rather about whether it has learned something correctly or incorrectly, in the form of rewards and punishments. This is called *Reinforcement Learning* and is the type of learning best performed in dynamic environments such as when driving a vehicle or playing a game against an opponent (Bishop, 2006).

Figure 55.1 summarizes different types of learning and how they relate to their subtasks.

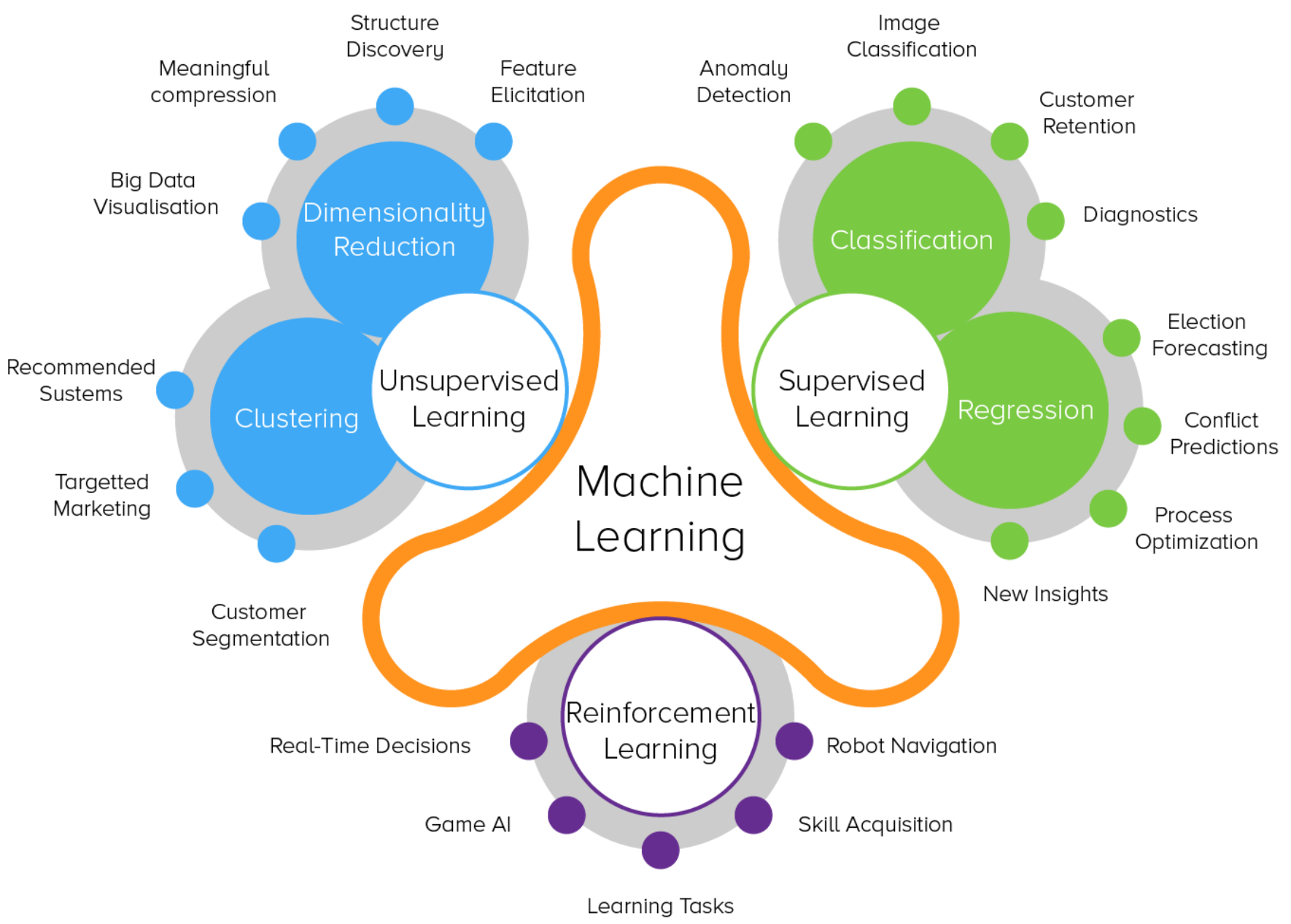

One of the most fruitful areas of machine learning applications in political science relates to work that treats text as data. Such quantitative text analysis could involve the following tasks: Assign a category to a group of documents or other elements (‘*classification*’): this is useful when, for example, there is a need to understand audience sentiment from social media or customer reviews or sort party manifestos into predefined categories on the ideological spectrum. Spam filtering is an example of classification from our contemporary daily life, where the inputs are email (or other) messages and the classes are ‘spam’ and ‘not spam’. The task involves a dataset containing text documents with labels, which is then used to train a classifier aiming to automatically classify the text documents into one or more predefined categories. Inputs are divided into two or more classes, and the algorithm assigns unseen inputs to one or more (multi-label classification) of these classes. This is typically tackled via supervised learning. In political science work, such models have been used, for example, to understand US Supreme Court decisions (Evans et al., 2007), party affiliation (Yu et al., 2008), and in measuring polarization (Peterson and Spirling, 2018).

Separate elements into groups (‘*clustering*’): this is similar to classification, only the groups are not known beforehand, hence this task usually involves unsupervised learning. Sanders et al. (2017) and Preoţiuc-Pietro et al. (2017) are examples of the potential use of clustering to better understand political ideologies and parliamentary topics.

Reduce the complexity of data: *dimensionality reduction* simplifies inputs by mapping them into a lower-dimensional space. Principal-components analysis and related methods like correspondence analysis have been used to analyze preferences for foreign aid (Baker, 2015) and the ideological mapping of candidates and campaign contributors (Bonica, 2014). Topic modelling is a related problem, where multiple documents are reduced to a smaller set of underlying themes or topics. Feature extraction is a type of dimensionality reduction task and can be accomplished using either semi-supervised or unsupervised learning. Selection and extraction of text features from documents or words is essential for text mining and information retrieval, where learning is done by seeking to reduce the dimension of the learning set into a set of features (Uysal, 2016; Nguyen et al., 2015).

Perform structured predictions*: structured prediction* or *structured (output) learning* is an umbrella term for supervised machine learning techniques that involve predicting structured objects, rather than scalar discrete or real values (BakIr, 2007). In Lafferty et al. (2001), for example, the issue of translating a natural-language sentence into a syntactic representation such as a parse tree can be seen as a structured-prediction problem in which the structured-output domain is the set of all possible parse trees.

The table below summarizes some of these techniques:

| Method | Type of learning | Examples |
|---|---|---|
| Classification | Supervised | • understand audience sentiment from social media<br>• sort party manifestos into predefined categories on the ideological spectrum<br>• understand US Supreme Court decisions (Evans et al., 2007),<br>• extract party affiliation (Yu et al., 2008),<br>• measure polarization (Peterson and Spirling, 2018) |
| Clustering | Unsupervised | • understand political ideologies and parliamentary topics (Sanders et al., 2017; Preoţiuc-Pietro et al., 2017) |
| Dimensionality Reduction e.g. Topic modelling, Feature Extraction | Semi-supervised Unsupervised | • preferences for foreign aid (Baker, 2015)<br>• ideological mapping of candidates and campaign contributors (Bonica, 2014)<br>• extraction of text features from documents (Uysal, 2016; Nguyen et al., 2015) |

*Table 55.1Overview of machine learning methods and examples from political science*

These political text-as-data applications are related to the broader field of NLP, which is concerned with the interactions between computers and human or natural languages (rather than formal languages). After the 1980s and alongside the developments in machine learning and advances in hardware and technology, NLP has mostly evolved around the use of statistical models to automatically identify patterns and structures in language, through the analysis of large sets of annotated texts or corpora. In addition to document classification and dimensionality-reduction applications in political science, leveraging the latest developments in machine learning and deep learning methods, the NLP field has made significant progress on several additional tasks:

- *Extracting text from an image.* Such a task usually involves a form of *Optical Character Recognition*, which can help with determining the corresponding text characters from an image of printed or handwritten text.

- *Identifying boundaries and segment text into smaller units (for example from documents to characters).* Examples of such tasks include morphological segmentation, word segmentation, and sentence-boundary disambiguation.

  *Morphological segmentation* is the field of separating words into individual morphemes and identifying the class of the morphemes is an essential step of text pre-processing before textual data can be used as an input in some machine learning algorithms. Some such tasks can be quite challenging to perform automatically, sometimes depending on morphological complexity (i.e. the internal structure of words) of the language being considered.

  *Word segmentation* or *tokenization* makes possible the separation of continuous text into separate words.

  *Sentence-boundary disambiguation* helps identify where a sentence starts and where it ends. This is not as simple as identifying where a period or other punctuation mark is, since not all punctuation signals the end of a sentence (consider abbreviations, for example) and not all sentences have punctuation.

*Assigning meaning to units. Part-of-speech tagging*, involves automatically determining and assigning a part of speech (e.g., a verb or a noun) to a word is usually the first step to looking at word context and meaning. Of course many words have more than one meaning or could be assigned different parts of speech, which can prove challenging for NLP, as it needs to select the meaning which makes more sense in the current context. With the emergence of deep learning methods, word embeddings have been used to capture semantic properties of words and their context (see the next section for a more detailed presentation).

*Extracting information from the text and synthesizing it.* NLP tasks such as *Named Entity Recognition, Sentiment Analysis, Machine Translation* and *Automated Text Summarization* build on the above tasks in order to identify and extract specific content from texts and synthesize it to generate new insights or content.

> *Machine Translation* studies ways to automate the translation between languages. Deep learning methods are improving the accuracy of algorithms for this task (Nallapati et al., 2016). This leads to scaling-up opportunities in comparative politics research (de Vries et al., 2018).
>
> *Named Entity Recognition* helps determine the elements in a text that are proper names (such as people or places) and what type of elements they are (for example, person, location, organization, etc.).
>
> *Sentiment Analysis* is the automatic extraction of opinions or subjective information from a set of documents or reviews, to determine 'polarity' about specific ideas. For example, scholars have used *Sentiment Analysis* to identify trends of public opinion in social media (Ceron et al., 2014; Proksch et al., 2015).
>
> *Automated Text Summarization* is a common dimensionality-reduction task in machine learning and NLP. It involves producing a readable, coherent, and fluent summary of a longer text, which should include the main points outlined in the document. *Extractive summarization* involves techniques such as identifying key words from the source document and combining them into a continuous text to make a summary. *Abstractive summarization* involves automatically paraphrasing or shortening parts of the original text.

With the deep learning methods being extremely data hungry, we believe that a primary area where the field will benefit from the latest technology is in the text-as-data or broader NLP domain. In what follows, we outline several deep learning models that have made recent advances in NLP possible and highlight how they can be used in political science research.

# Deep Learning NLP for Political Analysis

## Understanding 'Learning'

To define *deep learning* and understand the difference between deep learning and other machine learning approaches, first we need some idea of what machine learning algorithms *do*. As mentioned above, the field of machine learning is concerned with the question of how to construct computer programs that automatically improve with experience.

But what does *learning* mean in this context?

> A computer program is said to learn from experience *E* with respect to some class of tasks *T* and performance measure *P if its performance at tasks in T, as measured by P, improves with experience E*. (Mitchell, 1997; our emphasis)

This type of learning that particularly pertains to NLP regardless of the type of learning (supervised, unsupervised, active, etc.) is very much based on a 'bag-of-words' approach that only considers one dimension of the text, without taking onboard any of the contextual information – a rather 'shallow' type of learning.

Deep learning, on the other hand, offers the potential to combine multiple *layers* of representation of information, sometimes grouped in a hierarchical way.

## Understanding 'Deep'

Deep learning is a type of machine learning (representation learning) that enables a machine to automatically learn the patterns needed to perform regression or classification when provided with raw data. The approach puts an emphasis on learning successive *layers* of increasingly meaningful representations. It involves multiple levels of representation. Deng (2014: 199–200) define deep learning as a class of machine learning algorithms that use a cascade of multiple layers of nonlinear processing units for feature extraction and transformation, and each successive layer uses the output from the previous layer as input;

learn in supervised (e.g., classification) and/or unsupervised (e.g., pattern analysis) manners;

learn multiple levels of representations that correspond to different levels of abstraction – the levels form a hierarchy of concepts.

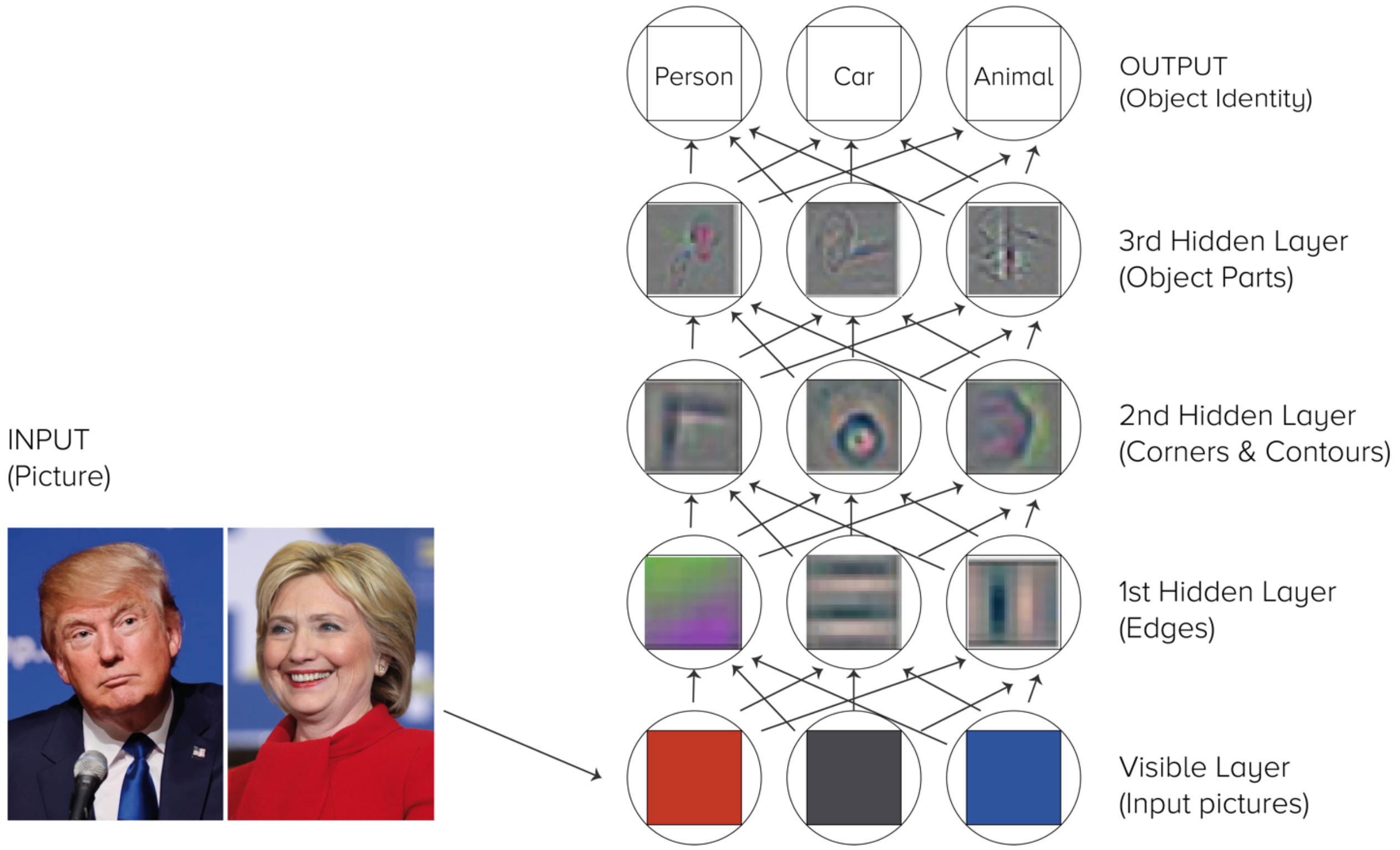

In deep learning, each level learns to transform its input data into a slightly more abstract and composite representation. In an image-recognition application, the raw input may be a matrix of pixels, the first representational layer may abstract the pixels and encode edges, the second layer may compose and encode the arrangements of edges, the third layer may encode eyes and a nose, and the fourth layer may recognize that the image contains a face (for more information about feature visualizations from computer-vision deep neural networks, see Olah et al., 2017 and Zhang and Zhu, 2018). Importantly, a deep learning process can learn which features to optimally place in which level *on its own*. Figure 55.2 shows how a deep learning hierarchy of complex concepts can be built from simpler concepts.

We will next discuss the application of deep learning algorithms in generating insights from images and text data.

## Working with Image Data

Convolutional neural networks (CNNs) are a category of artificial neural networks that have proven very effective when trying to classify or detect features in images. CNNs have been very successful at identifying objects, faces, and traffic signs in images and are currently advancing computer vision in robotics and self-driving vehicles.

CNNs have been trained on satellite imagery to map and estimate poverty, where data on economic livelihoods are scarce and where outcomes cannot be studied via other data. Jean et al. (2016) combine satellite imagery with survey data from five African countries (Nigeria, Tanzania, Uganda, Malawi, and Rwanda) to train a CNN to identify image features that can explain up to 75% of the variation in the local-level economic outcomes by estimating consumption expenditure. Figure 55.3 shows four different convolutional filters used for extracting these features, which identify (from left to right) features corresponding to urban areas, non-urban areas, water and roads. Babenko et al. (2017) focus on an urban subsample of satellite images in Mexico (using images from Digital Globe and Planet) identifying rural and urban 'pockets' of poverty that are inaccessible and changing frequently – areas that are unlikely to integrate without the support of the necessary policy measures (Figure 55.4).

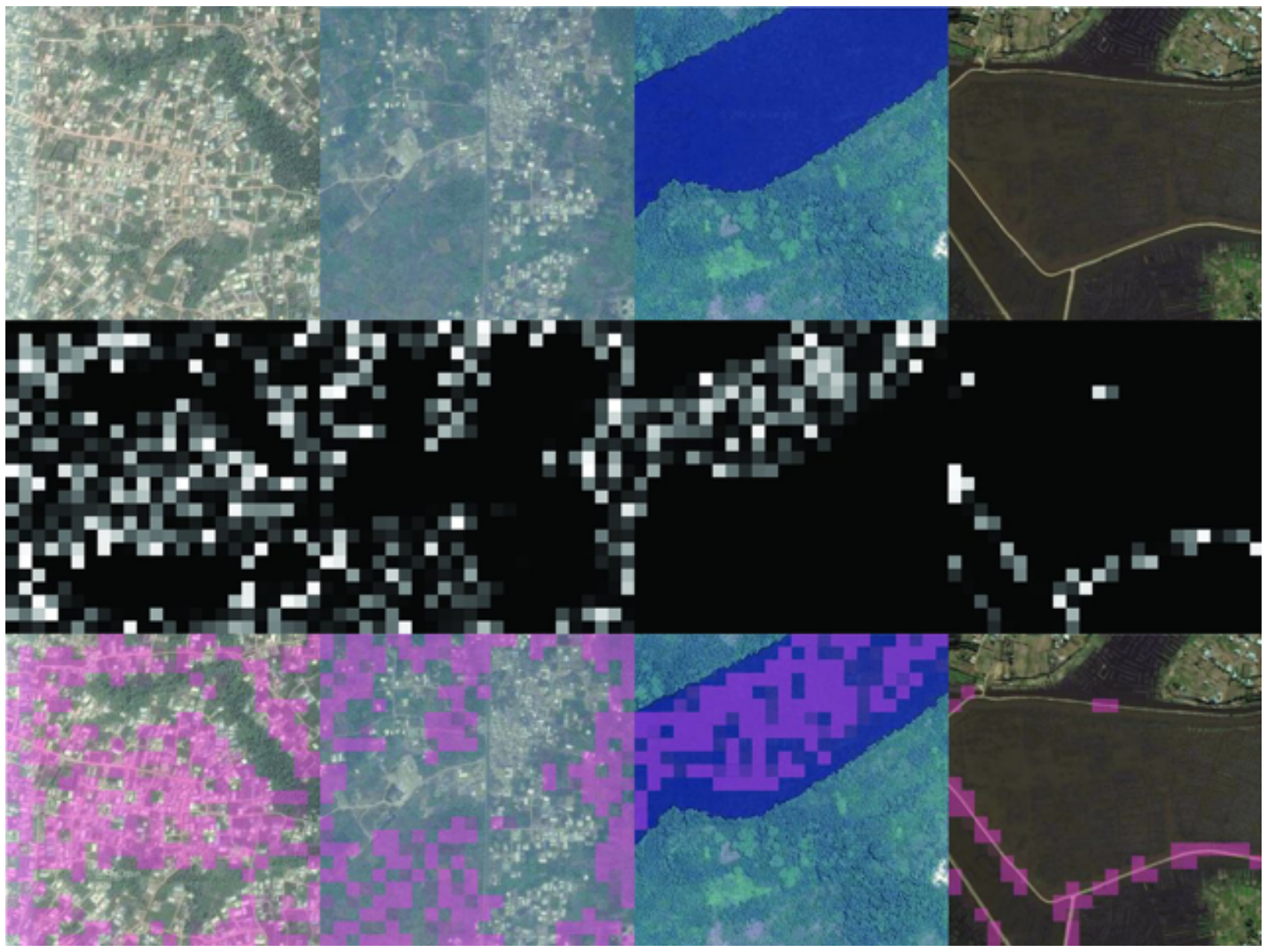

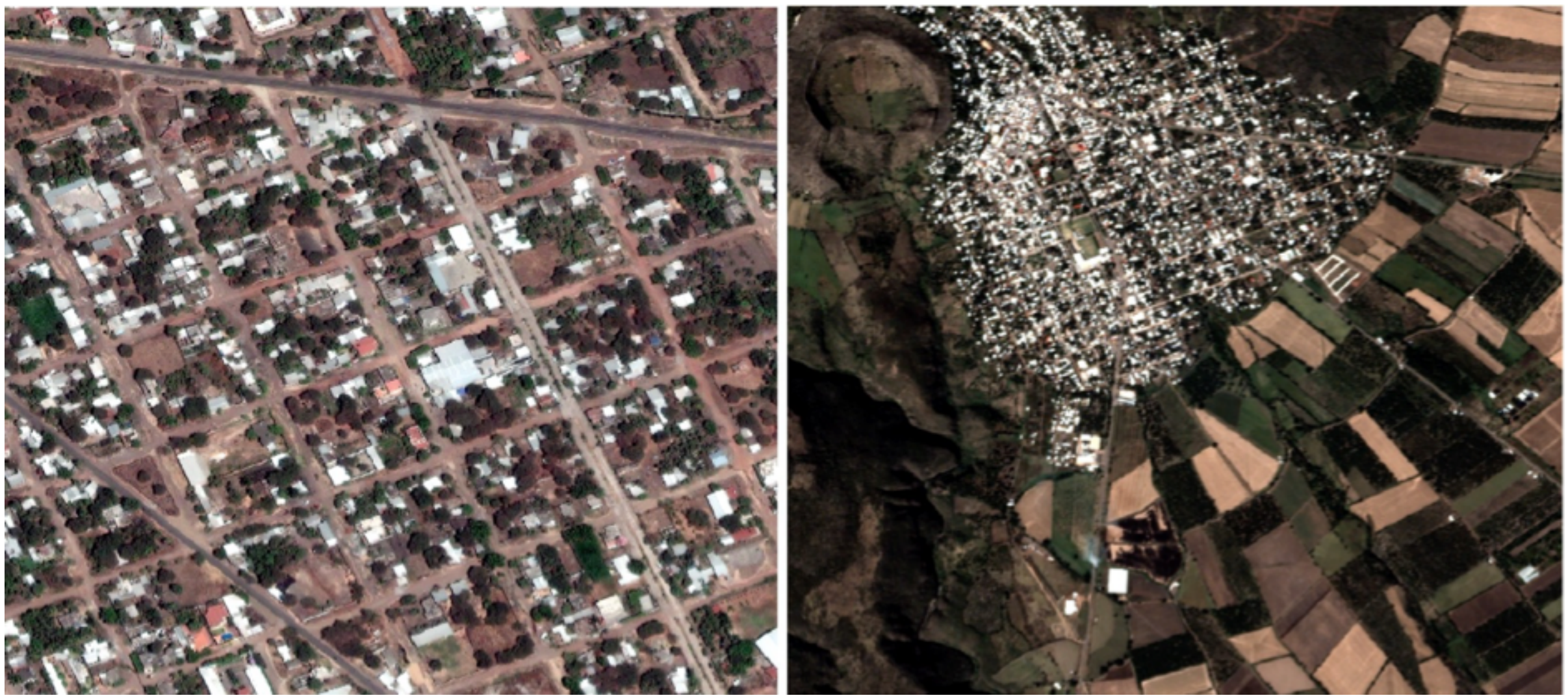

CNNs have also been used to map informal settlements ('slums') in developing countries, using high- and low-resolution satellite imagery (Helber et al., 2018), to help international aid organizations to provide effective social and economic aid.

But how do they work?

Analogous to how children learn to recognize a cat from a dog, we need to 'show' an algorithm millions of pictures ('input') of a dog before it can reliably make generalizations and predictions for images it has never seen before. However, machines do not 'see' in the same way we do – their ‘language’ consists of numbers. One way around this is to represent every image as multi-dimensional arrays of numbers, and CNNs offer a way to move from an image to a set of vectors.

The main building block of CNN is the *convolutional layer, filter, or kernel*. Convolution is a mathematical operation that allows us to condense information by combining two functions into one. Take the very simple, pixelated representation of a black and white heart in Figure 55.5 element (a) for example. If each cell is a pixel, then we could represent black pixels with value 1 and white pixels with value 0 (see Figure 55.5, element (b)) – this is the ‘input’.

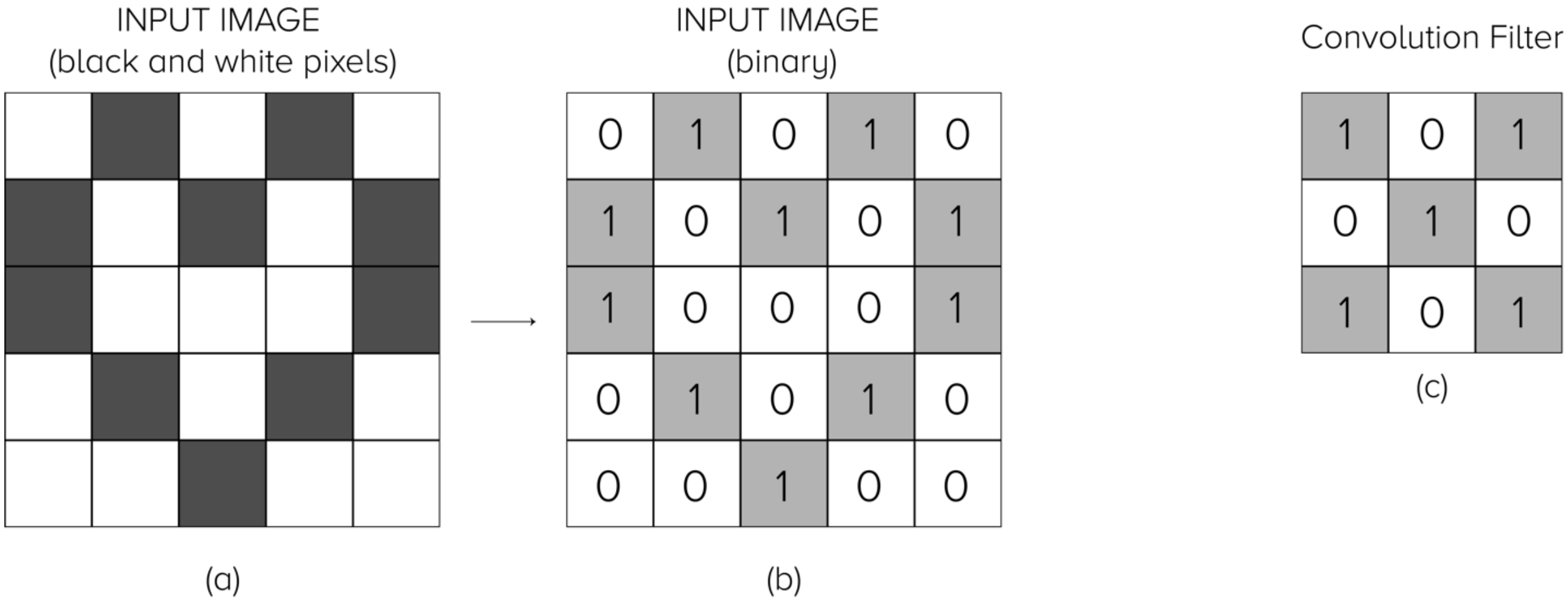

Using a filter, as in Figure 55.5 element (c), with predefined black and white pixels, we can now perform a convolution and create a ‘feature map’ (Figure 55.6, element (d)) by layering the filter on top of the input and sliding it for each row. At every step, we perform element-wise matrix multiplication and sum the result, which goes into the feature map – represented in the black background in Figure 55.6.

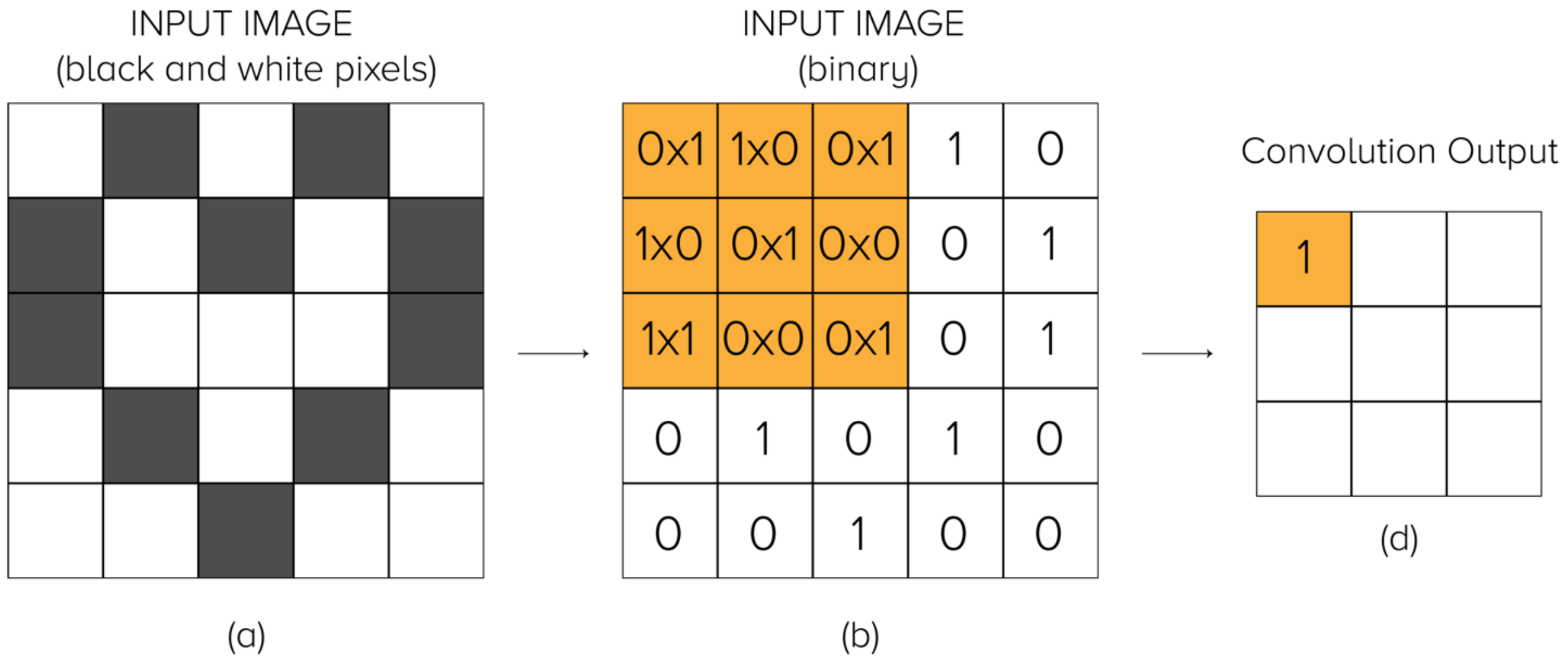

We then slide the filter over the next position and perform the same multiplication (see Figure 55.7).

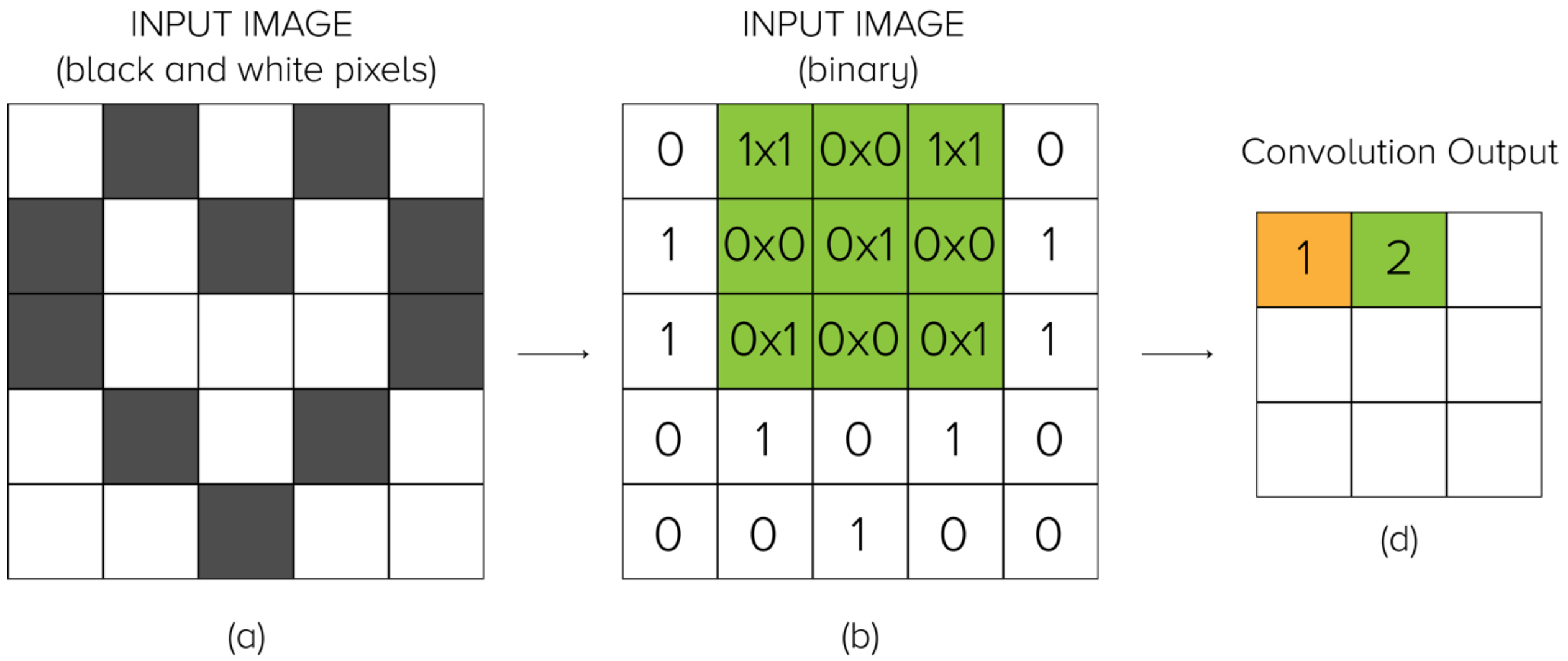

We do the same until the ‘input’ is reduced from a 5x5 matrix to a 3x3 feature map.

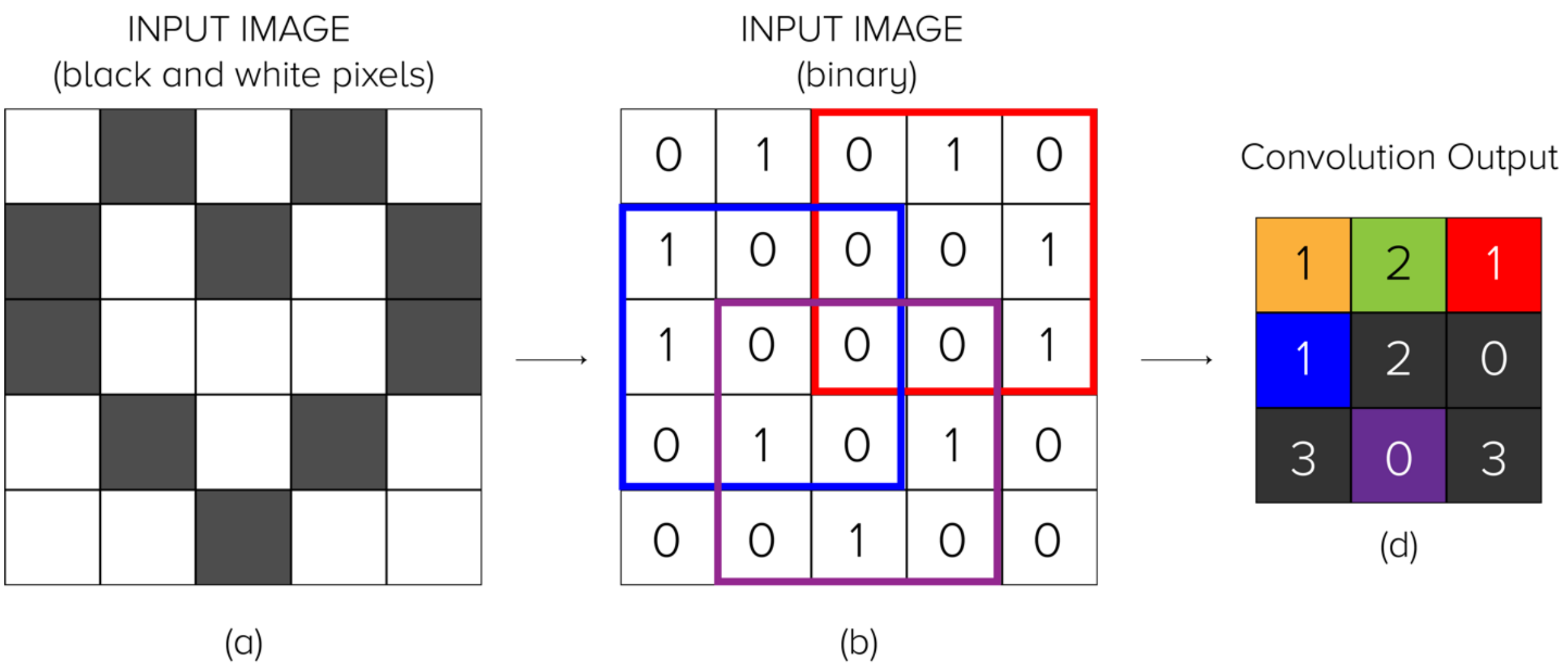

We repeat until the 'input' is reduced from a 5x5 matrix to a 3x3 feature map, as in Figure 55.8 element (c) above. The example above is a two-dimensional convolution using a 3x3 filter – in reality, these convolutions are performed in three dimensions (width, height, and RGB color channel) with the filter being 3D as well. Multiple convolutions take place on an input, each using a different filter with a distinct feature map as the output. After a convolution operation, we usually perform *pooling* (usually *max pooling*, i.e. taking the max value in the pooling window) to reduce the dimensionality and reduce the number of parameters (see Figure 55.9).

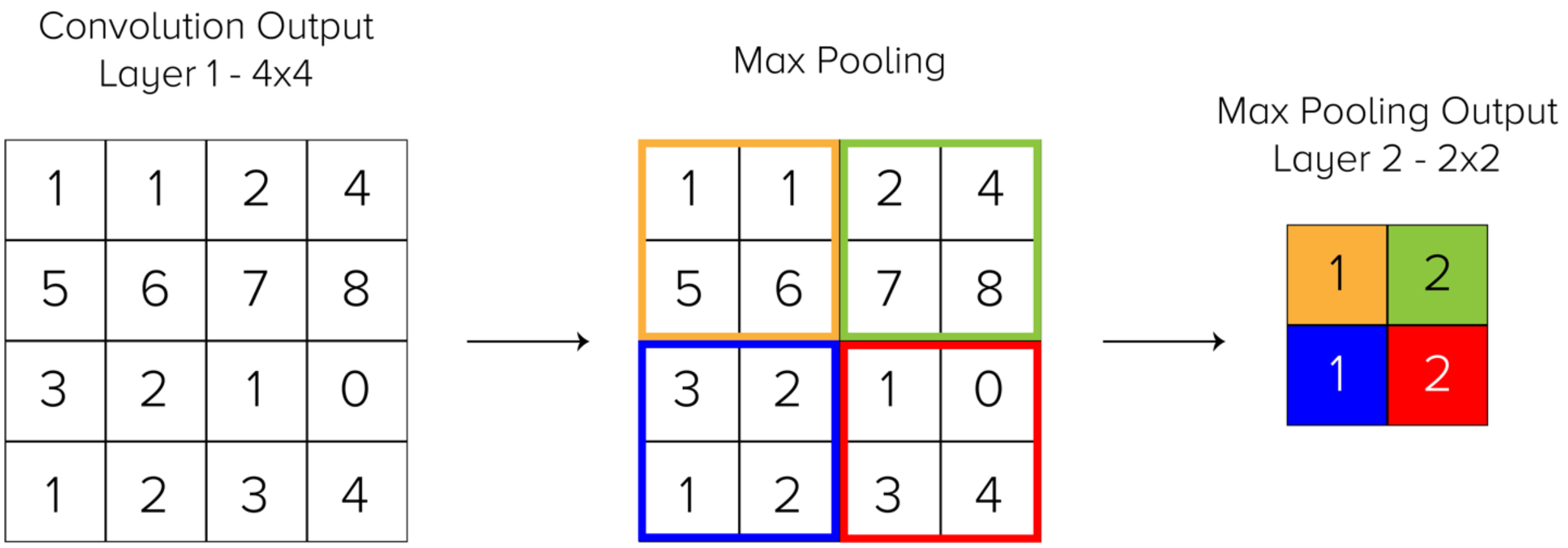

This is crucial when dealing with the volume of data that is fed to the algorithm, as it both speeds training time and helps avoid overfitting of the algorithm.

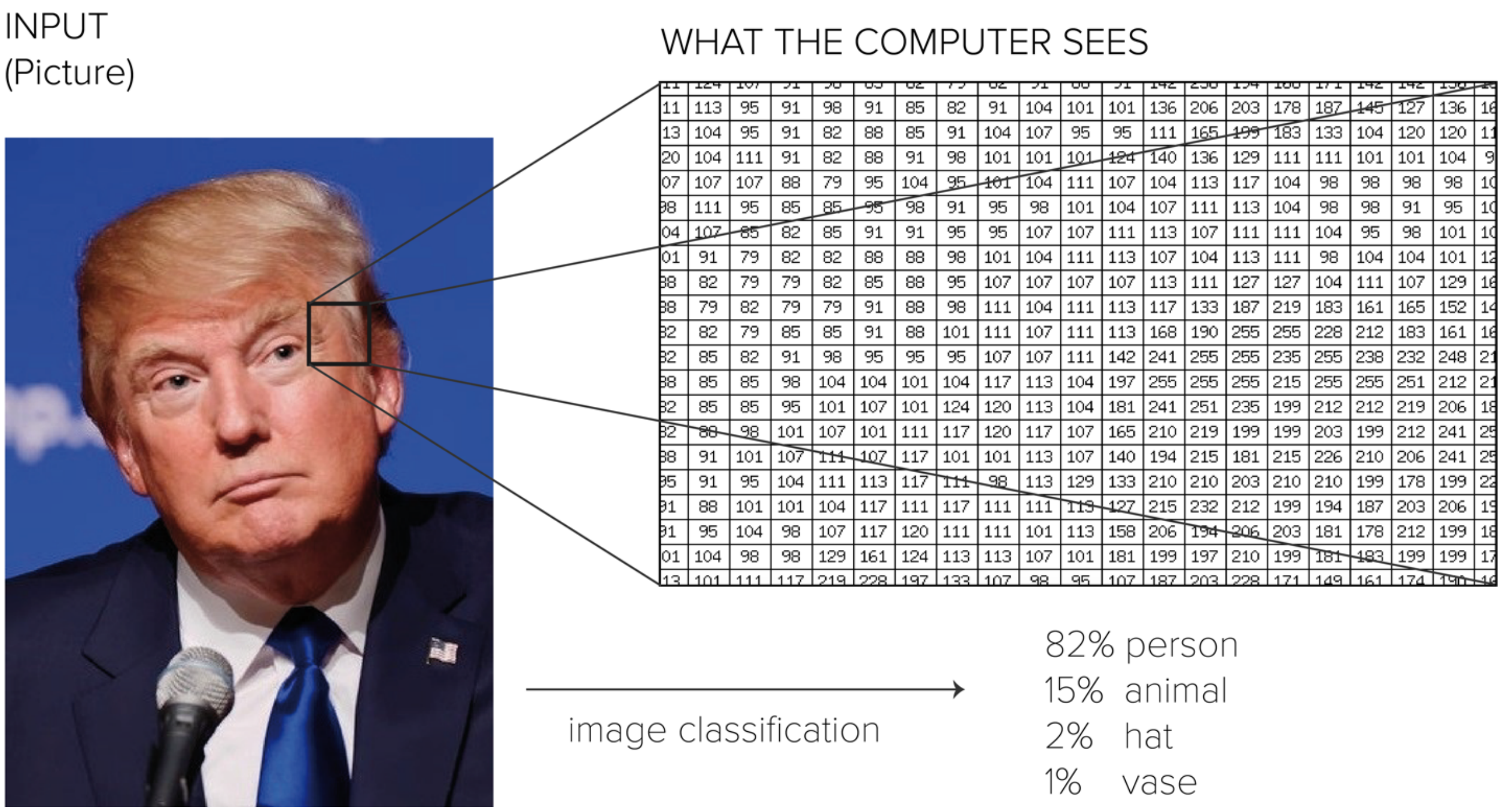

CNNs seem to suit the task of image classification, as they can help us predict a distribution over specific labels (as in Figure 55.10) to indicate confidence of prediction for a given image. But what about text data?

## Working with Text Data

The study of political discourse using text as data has a long tradition in political science. Political texts have long been used as an important form of social practice that contributes to the construction of social identities and relations (Fairclough, 1989, 1992; Laclau and Mouffe, 1985). Text as a representation of discourses has been studied systematically to derive information about actors and combine them with additional resources such as surveys and observations, as well as knowledge and reflective understanding of the context by scholars, yet not in a reproducible and quantifiable way (see Blommaert and Bulcaen, 2000, for a review).

Over the past two decades, scholars have sought to extract information such as policy and ideology positions and gauge citizen political engagement by treating words as data in a more consistent way. Since some of the earliest implementations of text-scaling methods such as *Wordscores* (Laver et al., 2003) and *Wordfish* (Slapin and Proksch, 2008) to estimate party positions from texts and the increasing availability of annotated political corpora, the availability and complexity of quantitative text-analysis methods have increased dramatically (Barberá, 2015; Grimmer and Stewart, 2013; Herzog and Benoit, 2015; Lauderdale and Herzog, 2016). Most of these methods tend to involve a 'bag-of-words' approach to determine relevance and cluster documents or their parts in groups (see also Laver, 2014). Such approaches assume that each document can be represented by a multiset ('bag') of its words, that ignores word order and grammar. Word frequencies in the document are then used to classify the document into a category. Some methods like Wordscores employ a version of the Naive Bayes classifier (Benoit and Nulty, 2013) in a supervised learning setting by leveraging pre-labelled training data, whereas others, like WordFish, are based on a Poisson distribution of word frequencies, with ideological positions estimated using an expectation-maximization algorithm (Proksch and Slapin, 2009; Slapin and Proksch, 2008).

What these approaches do not capture, though, is the linguistic and semiological context, i.e. the information provided by the words around the target elements. Such a context would allow for a better representation of that context and offer a richer understanding of word relationships in a political text. One way to do that is by using *word embeddings*, a set of methods to model language, combining concepts from NLP and graph theory.

## Representing Words in Context: Word Embeddings

Word embeddings are a set of language modelling and dimensionality-reduction techniques, where words or phrases from a document are mapped to vectors or numbers. They usually involve a mathematical embedding from a space with a single dimension for each word to a continuous vector space with a reduced dimension. The underlying idea is that '[y]ou shall know a word by the company it keeps' (Firth, 1957: 11), and it has evolved from ideas in structuralist linguistics and ordinary language philosophy, as expressed in the work of Zelling Harris, John Firth, Ludwig Wittgenstein, and vector-space models for information retrieval in the late 1960s to the 1980s. In the 2000s, Bengio et al. (2006) and Holmes and Jain (2006) provided a series of papers on the 'Neural Probabilistic Language Models' in order to address the issues of dimensionality of word representations in contexts, by facilitating learning of a 'distributed representation of words'. The method developed gradually and really took off after 2010, partly due to major advances in the quality of vectors and the training speeds of the models.

There are many variations of word-embedding implementations, and many research groups have created similar but slightly different types of word embeddings that can be used in the deep learning pipelines. Popular implementations include Google's Word2Vec (Mikolov et al., 2013), Stanford University's GloVe (Pennington et al., 2014), and Facebook's fastText (Bojanowski et al., 2016). For a recent discussion of word embeddings in a political science context, see Spirling and Rodriguez (2019).

Now that we have a mechanism to turn text into dense vectors (very much like we did with the image of the heart in the previous section), let's see how CNNs can be applied to NLP tasks for political texts.

## CNNs for Text Analysis

CNNs have recently been applied to various NLP tasks with very good results in accuracy and precision (Johnson and Zhang, 2014; Kalchbrenner et al., 2014; Kim, 2014).

Instead of image pixels, each row of the matrix corresponds to one token (usually a word, but it could also be a character; see Jacovi et al., 2018 and Zhang et al., 2015) or rather a vector that represents a word. These vectors are typically *word embeddings* such as Word2Vec or GloVe (see previous section). Kim (2014) describes the general approach of using CNNs for NLP, assuming a single layer of networks and pretrained static word vectors on very large corpora (Word2Vec vectors from Google, trained on 100 billion tokens from Google News). Sentences are mapped to embedding vectors and are available as a matrix input to the model. Convolutions are performed across the input word-wise using differently sized kernels, such as two or three words at a time. The resulting feature maps are then processed using a max pooling layer to condense or summarize the extracted features. Figure 55.11 shows a single-layer CNN architecture for sentence classification from Kim (2014).

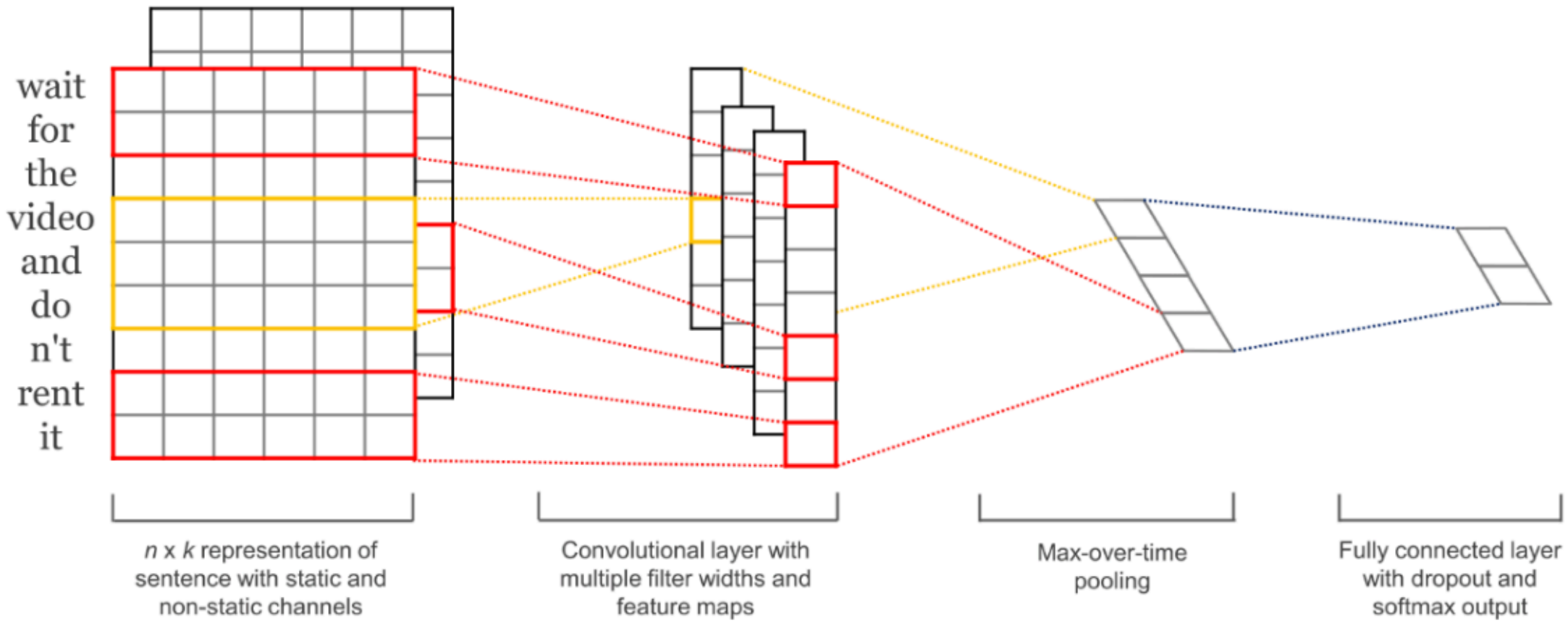

*Figure 11. Illustration of a single-layer Convolutional Neural Network (CNN) architecture for sentence classification from (Kim, 2014))*

Figure 55.12 shows how a CNN would work for a sentence-classification task adapted from Zhang and Wallace (2015). Assuming the sentence we wanted to classify was Michelle Obama's *'When they go low, we go high'*, this would generate a 7x4 sentence matrix, with three filter region sizes: 2, 3, and 4, each of which has two filters for each region size. Every filter performs convolution on the sentence matrix and generates (variable-length) feature maps. Then, 1-max pooling is performed over each map, i.e. the largest number from each feature map is recorded. Thus, a univariate feature vector is generated from all six maps, and these six features are concatenated to form a feature vector for the penultimate layer. The final *softmax* layer then receives this feature vector as input and uses it to classify the sentence; here, we assume binary classification and hence depict two possible output states.

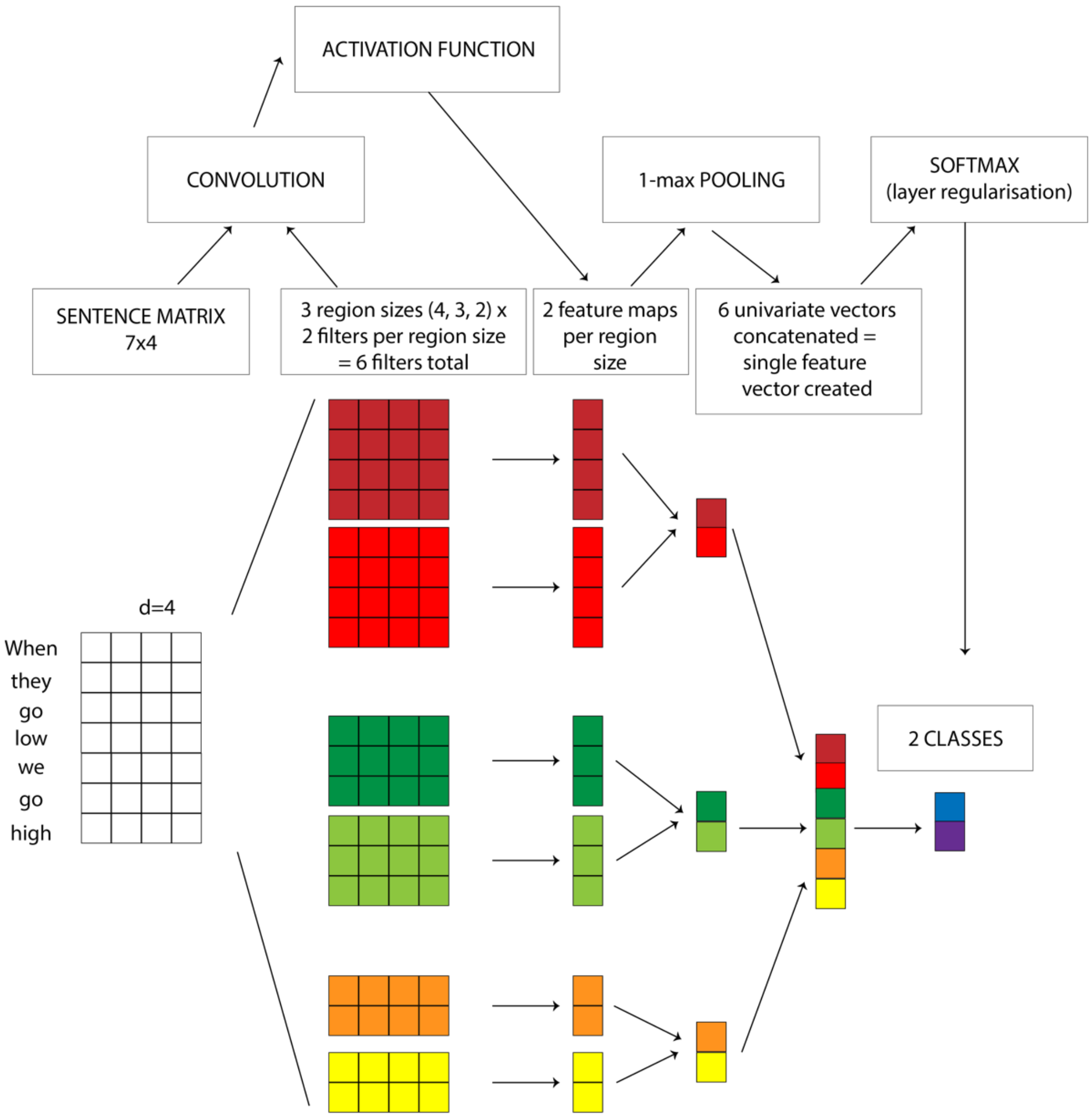

Despite CNNs being a little unintuitive in their language implementation, they perform really well on tasks like text classification. They are very fast, as convolutions are highly parallelizable, form an integral part of computer graphics, and are implemented on graphical processing units (GPUs). They also work much better compared to other 'bag-of-words' approaches such as n-grams, as they can learn representations automatically without the need to represent the whole vocabulary (whereas in the case of n-grams, for example, if we had a large vocabulary, computing anything beyond tri-grams would become quite expensive in terms of computational power), with architectures as deep as 29 layers performing sufficiently well (Zhang et al., 2015).

CNNs have been successfully deployed for NLP tasks such as automatic summarization, fake news detection and text classification. Narayan et al. (2018), for example, apply CNNs to automatically summarize a real-world, large-scale dataset of online articles from the British Broadcasting Corporation (BBC). They demonstrate experimentally that this architecture captures long-range relationships in a document and recognizes related content, outperforming other state-of-the-art abstractive approaches when evaluated automatically and by humans.

Yamshchikov and Rezagholi (2018) develop a model of binary text classifiers based on CNNs, which helps them label statements in the political programs of the Democratic and Republican parties in the United States, whereas Bilbao-Jayo and Almeida (2018) propose a new approach to automate the analysis of texts in the Manifestos Project, to allow for a quicker and more streamlined classification of such types of political texts.

The Manifesto Project (Lehmann et al., 2018) includes data on parties' policy positions, derived from content analysis of parties' electoral manifestos. It covers over 1,000 parties from 1945 until today in over 50 countries on five continents. The corpus includes manually annotated election manifestos using the Manifesto Project coding scheme, which is widely used in comparative politics research. Bilbao-Jayo and Almeida (2018) use multi-scale CNNs with word embeddings and two types of context data as extra features, like the previous sentence in the manifesto and the political party. Their model achieves reasonably high performance of the classifier across several languages of the Manifesto Project.

Another type of neural network that has shown good performance in NLP tasks are recurrent neural networks (RNNs) and, in particular, a variation of that algorithm, the long short-term memory (LSTM) RNNs.

## LSTM RNNs for Text Analysis

As you read this paragraph, you understand each word based on your understanding of previous words – those right before this word, words expressed in the paragraphs and sections above, as well as words that you might have read in the previous chapters of this *Handbook* (or even words that you have read in other books and articles).

Every time we read a new word, we do not forget what we read before – our understanding has some degree of *persistence*. Unfortunately, CNNs cannot reason about previous steps in the learning process to inform later ones. RNNs overcome this issue because they permit loops, thus allowing for the information in the neural network to persist. A simple RNN is a class of artificial neural networks where connections between nodes form a directed graph along a sequence, incorporating previous knowledge (see Figure 55.13, adapted from Olah, 2015).

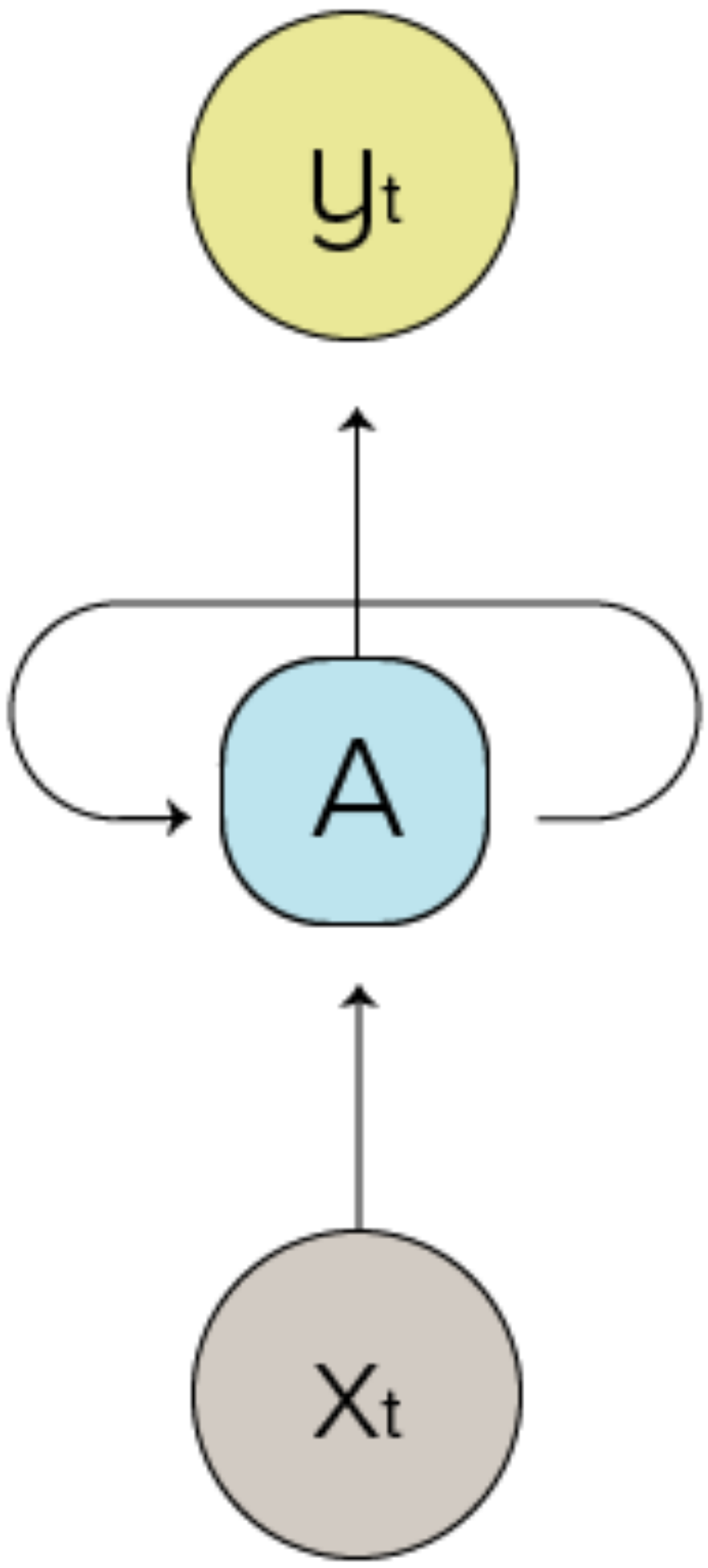
yt
A
Xt

A sequence of RNN blocks can be regarded as multiple copies of the same network, linked to one another like a chain, each passing an input to its future self (Figure 55.14). This enables it to display dynamic temporal behavior for a time sequence and make these networks work really robustly with sequence data such as text, time-series data, videos, and even DNA sequences.

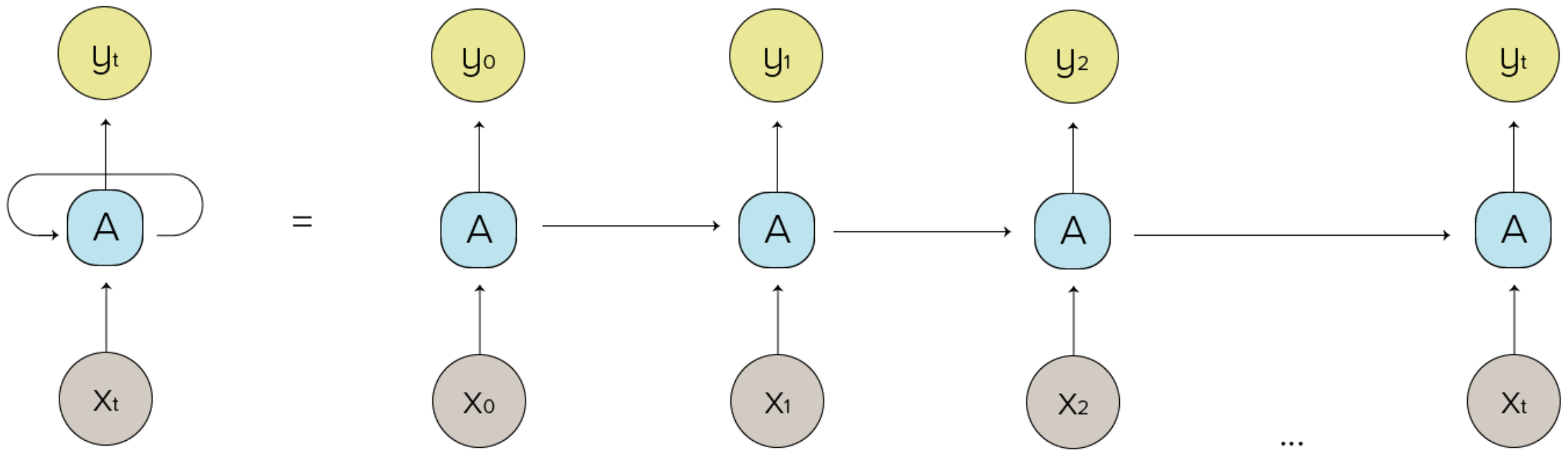

This suits textual data, which for the most part is sequence or list data, and which has been applied with success to NLP tasks such as speech recognition, language modelling, translation, and image captioning (Ba et al., 2014; Gregor et al., 2015). However, simple RNNs are not well suited for remembering information that is not close to the current node they are in (also called long-distance dependencies), a problem detailed in Bengio et al. (1994).

LSTM neural networks (Hochreiter and Schmidhuber, 1997) provide a solution to this issue. LSTMs also have the RNN chain-like structure, but the repeating module has a different structure. Instead of having a single neural network layer, there are four, all interacting in a special way. Figure 55.15 shows the repeating module in a standard RNN with a single layer (A1) and an LSTM with 4 interacting layers (A2). The LSTM has the advantage of incorporating context from both the input (x) and the previous knowledge (represented with dashed lines in A2) and also feed the augmented knowledge to the next iteration.

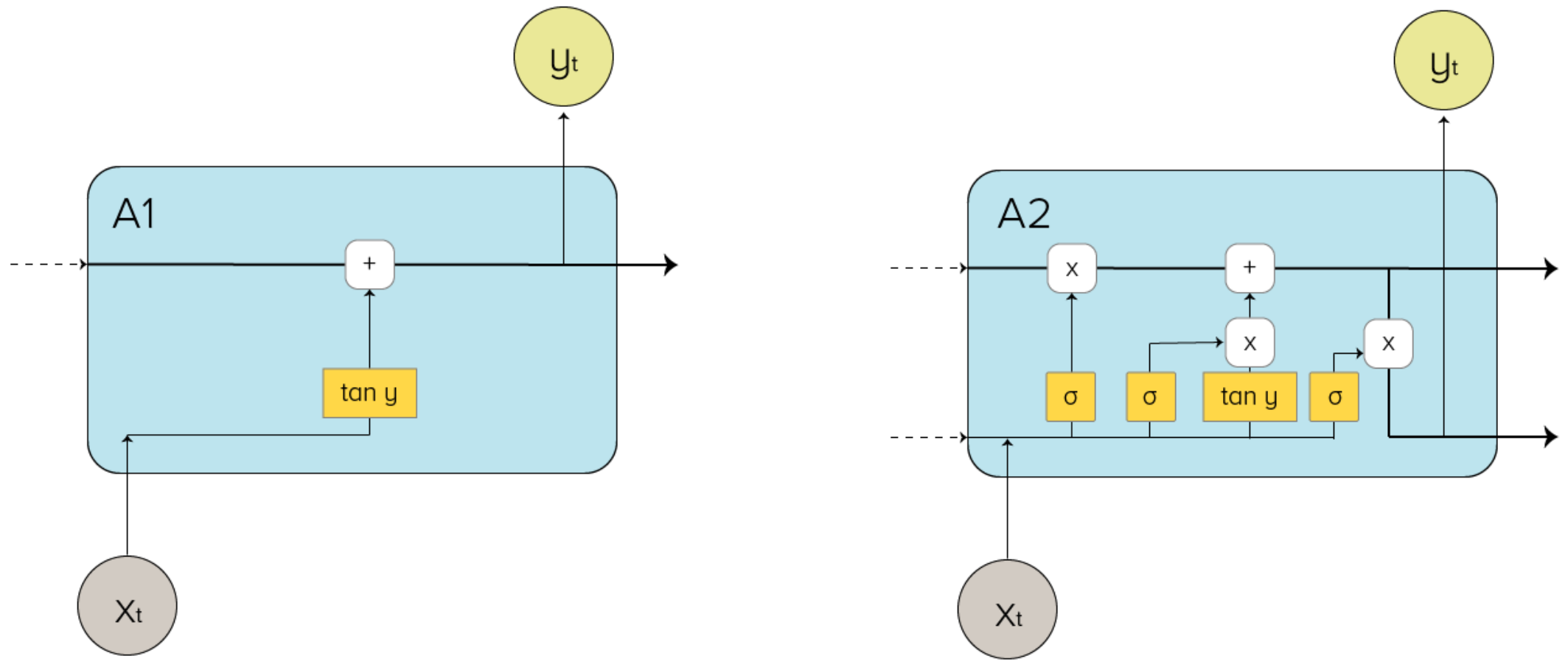

Standard LSTMs (like those in Figure 55.15) are *unidirectional* – in other words, they preserve information from the past inputs that have already passed through the different iterations of the hidden layers of the neural network. Take for example the following word sequence:

‘Let’s make …’

There are a lot of possibilities for what word sequences could follow. All the sentences below are possible:

‘Let’s make some cake!’

‘Let’s make fun of Bob!’

‘Let’s make my friend see some sense, because I think she is making a huge mistake!’

What if you knew that the words that followed the first word sequence were actually these?

“Let’s make … great again!”

Now the range of options is narrower, and it is easy to predict that the next word is probably a noun phrase such as ‘America’ or ‘this business’.

A unidirectional LSTM will only be able to consider past input ('let's make'). If you wish to see the future, you would need to use a *bidirectional* LSTM, which will run the input in two ways: one from the past to the future and one from the future to the past. When running backwards, it preserves information from the future, and by combining this knowledge with the past, it provides improved and more contextualized predictions.

Both types of LSTMs have been used to detect fake news and propaganda discourse in traditional and social media text, where the problem of detecting bots – automated social media accounts governed by software but disguised as human users – has strong societal and political implications.

Kudugunta and Ferrara (2018) propose a deep neural network based on contextual LSTM architecture, that exploits both content and metadata to detect bots at the tweet level. Their proposed technique is based on synthetic minority oversampling to generate a large labelled dataset suitable for deep nets training, from a minimal amount of labelled data (roughly 3,000 examples of sophisticated Twitter bots). The proposed model can, from the first tweet, achieve high classification accuracy (> 96%) in separating bots from humans.

Event detection using neural-network algorithms on tweets describing an event is another area of application of particular interest to media agencies and policy makers. Iyyer et al. (2014) assume that an individual's words often reveal their political ideology, and they use RNNs to identify the political position demonstrated at the sentence level, reporting that their model outperforms 'bag of words' or wordlists models in both the training and a newly annotated dataset. Makino et al. (2018), for example, propose a method to input and concatenate character and word sequences in Japanese tweets by using CNNs and reporting an improved accuracy score, whereas Rao and Spasojevic (2016) apply word embeddings and LSTM to text classification problems, where the classification criteria are decided by the context of the application. They show that using LSTMs with word embeddings vastly outperforms traditional techniques, particularly in the domain of text classification of social media messages' political leaning. The research reports an accuracy of classification of 87.57%, something that has been used in practice to help company agents provide customer support by prioritizing which messages to respond to.

Other scholars have used hybrid neural-network approaches to work with text, by combining aspects of the CNN and RNN algorithms. Ajao et al. (2018), for example, propose a framework that detects and classifies fake news messages from Twitter posts, using such a hybrid of CNNs and LSTM RNNs, an approach that allows them to identify relevant features associated with fake news stories without previous knowledge of the domain. Singh et al. (2018) use a combination of the CNN, LSTM, and bidirectional LSTM to detect (overt and covert) aggression and hate speech on Facebook and social media comments, where the rise of user-generated content in social media coupled with almost non-existent moderation in many such systems has seen aggressive content rise.

Hybrid neural-network approaches also perform well in the task of automatic identification and verification of political claims. The task assumes that given a debate or political speech, we can produce a ranked list of all of the sentences based on their worthiness for fact checking – potential uses of this would be to predict which claims in a debate should be prioritized for fact-checking. As outlined in Atanasova et al. (2018), of a total of seven models compared, the most successful approaches used by the participants relied on recurrent and multi-layer neural networks, as well as combinations of distributional representations, matching claims' vocabulary against lexicons, and measures of syntactic dependency.

## Working with Multimodal Data

With the resurgence of deep learning for modeling data, the parallel progress in fields of computer vision and NLP, as well as with the increasing availability of text/image datasets, there has been a growing interest in using multimodal data that combines text with images. The popularity of crowd-sourcing tools for generating new, rich datasets combining visual and language content has been another important factor favoring multimodal input approaches.

Ramisa et al. (2018), for example, have compiled a large-scale dataset of news articles with rich metadata. The dataset, BreakingNews, consists of approximately 100,000 news articles collected over 2014, illustrated with one to three images and their corresponding captions. Each article is enriched with other data like related images from Google Images, tags, shallow and deep linguistic features (e.g., parts of speech, semantic topics, or outcomes of a sentiment analyzer), GPS latitude/longitude coordinates, and reader comments. The dataset is an excellent benchmark for taking joint vision and language developments a step further. Figure 55.16 illustrates the different components of the Ramisa et al. (2018) *BreakingNews* corpus, which contains a variety of news-related information for about 100K news articles. The figure shows two sample images. Such a volume of heterogeneous data makes *BreakingNews* a good benchmark for several tasks exploring the relation between text and images.

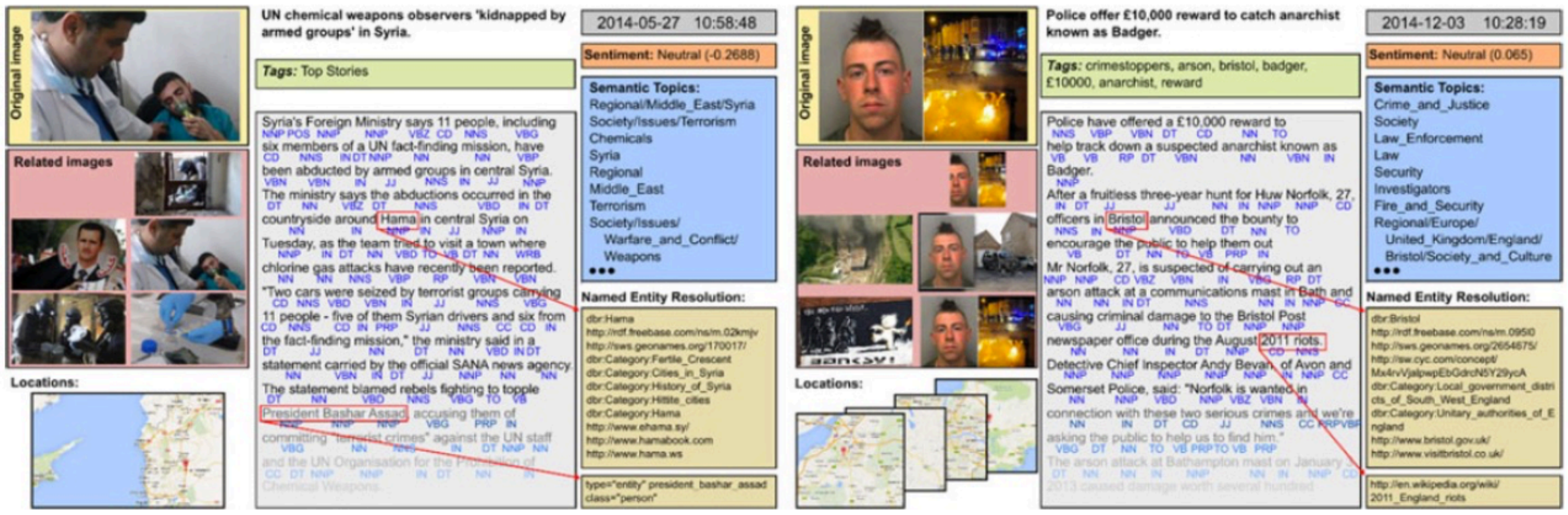

*Figure 17. The BreakingNews dataset (Ramisa, Yan, Moreno-Noguer, & Mikolajczyk, 2018). The dataset contains a variety of news-related information including: the text of the article, captions, related images, part-of-speech tagging, GPS coordinates, semantic topics list or results of sentiment analysis, for about 100K news articles. The figure shows two sample images. All this volume of heterogeneous data makes BreakingNews an appropriate benchmark for several tasks exploring the relation between text and images.*

The paper used CNN for source detection, geolocation prediction, and article illustration, and a mixed LSTM/CNNs model for caption generation. Overall results were very promising, especially for the tasks of source detection, article illustration, and geolocation. The automatic caption-generation task, however, demonstrated sensitivity to loosely related text and images.

Ajao et al. (2018) also fed mixed data inputs (text and images) to CNNs in order to detect fake news in political-debate speech, and they noted that except for the usual patterns in what would be considered misinformation, there also exists some hidden patterns in the words and images that can be captured with a set of latent features extracted via the multiple convolutional layers in the model. They put forward the TI-CNN (text and image information based convolutional neural network) model, whereby explicit and latent features can be projected into a unified feature space, with the TI-CNN able to be trained with both the text and image information simultaneously.

## Recent Developments

Deep neural networks have revolutionized the field of NLP. Furthermore, deep learning in NLP is undergoing an 'ImageNet' moment. In a paradigm shift, instead of using word embeddings as initializations of the first layer of the networks, we are now moving to pretraining the entire models that capture hierarchical representations and bring us closer to solving complex language-understanding tasks. When the ImageNet challenge AlexNet (Krizhevsky et al., 2012) solution showed a dramatically improved performance of deep learning models compared to traditional competitors, it arguably spurred the whole deep learning research wave. Over the last 18 months, pretrained language models have blown out of the water previous state-of-the-art results across many NLP tasks. These advances can be characterized within the broader framework of transfer learning, where the weights learned in state-of-the-art models can be used to initialize models for different datasets, and this 'fine-tuning' achieves superior performance even with as little as one positive example per category (Ruder et al., 2019).

One of the assumptions of standard word embeddings like Word2Vec is that the meaning of the word is relatively stable across sentences. An alternative is to develop contextualized embeddings as part of the language models. Embeddings from language models (ELMo) (Peters et al., 2018), universal language model fine-tuning (ULMFiT) (Howard and Ruder, 2018), and generative pretraining transformer (OpenAI GPT) (Radford et al., 2018) were initial extremely successful pretrained language models.

More recently GPT2 (Radford et al. 2019) extended the previous GPT model and was used to generate realistic-sounding artificial text. Bullock and Luengo-Oroz (2019) used the pretrained GPT2 model to generate fake but natural-sounding speeches in the United Nations General Debate (see Baturo et al., 2017, for more details about the data and a substantive example). Bidirectional encoder representations from transformers (BERT) (Devlin et al., 2019) extended GPT through bi-directional training and dramatically improved performance on various metrics.

While BERT was the reigning champion for several months, it may have recently been overtaken by XLNet (Yang et al., 2019), which outperforms BERT on about 20 NLP tasks.

In parallel with the advances in transfer learning, we are also further understanding what we are learning with the deep neural networks. Liu et al. (2019) show that RNNs (and LSTMs in particular) pick up general linguistic properties, with the lowest layers representing morphology and being the most transferable between tasks, middle layers representing syntax, and the highest layers representing task-specific semantics. Large pretrained language models do not exhibit the same monotonic increase in task specificity, with the middle layers being the most transferrable. Tenney et al. (2019) focus on BERT and show that the model represents the steps of the traditional NLP pipeline, with the parts-of-speech tagging followed by parsing, named-entity recognition, semantic roles, and, finally, coreference. Furthermore, the model adjusts the pipeline dynamically, taking into account complex interactions between different levels of hierarchical information.

Detailed discussion of the above models is beyond the scope of this chapter. Instead, we want to emphasize the pace of development in NLP research, which is leveraging pretrained language models for downstream tasks. Instead of downloading pretrained word embeddings like Word2Vec or GloVe as discussed earlier in the chapter, we are now in a position to download pretrained language models and fine-tune them to a specific task.

# Conclusion

It is appealing to think of machine learning algorithms as objective, unbiased actors that are beyond the influence of human prejudices. It is also appealing to think of empirical research in political science that utilizes machine learning algorithms as being sufficiently removed from any potential bias. Unfortunately, this is rarely the case.

Algorithms are designed by humans and learn by observing patterns in the data that very often represent biased human behavior. It is no surprise that algorithms tend to adopt and, in some occasions, perpetuate and reinforce the experiences and predispositions of the humans that have constructed them and those of society as a whole; this is also known as *algorithmic bias*. Although machine learning has been transformative in many fields, it has received criticism in the areas of causal inference, algorithmic bias, and data privacy. This is forming into a distinct area of social science research, focusing on the lack of (suitable) training data, difficulties of data access and data sharing, data bias and data provenance, privacy preserving data usage, and inadequate tasks, tools and evaluation settings (Danks and London, 2017).

The quality of insights delivered by algorithms crucially depends on data quality and data provenance. In particular, in each case, we need to effectively query very distinct (heterogeneous) data sources before we can extract and transform them for input into the data models. Common aspects of data quality that may affect the robustness of insights include consistency, integrity, accuracy, and completeness. How image or textual data is pre-processed may affect how data is interpreted and may also lead to biases. For example, dataset biases in computer vision can lead to feature representation flaws where CNNs, despite high accuracy, learn from unreliable co-appearing contexts (Zhang et al., 2018).

The consequences of biased algorithms can be quite real and severe. In 2016, an investigative study by ProPublica (Angwin et al., 2016) provided evidence that a risk-assessment machine learning algorithm used by US courts wrongly flagged non-white defendants at almost twice the rate of white defendants. More recently, Wang and Kosinski (2018) showed how deep neural networks can outperform humans in detecting sexual orientation. Apart from the ethical issues of the study, the ease of deployment of such 'AI Gaydar' raises issues of people's privacy and safety.

The issues of algorithmic bias are also highlighted in the Wellcome Trust Report (Matthew Fenech et al., 2018) with a focus on how AI has been used for health research. The report identifies, among other ethical, social, and political challenges, issues around implications of algorithmic transparency and explainability on health, the difference between an algorithmic decision and a human decision, and what makes algorithms, and the entities that create them, trustworthy. The report highlights the importance of stakeholders across the public- and private-sector organizations collaborating in the development of AI technology, and it raises awareness of the need for AI to be regulated.

Such algorithmic-bias issues may seem to be removed from everyday political science research. However, various methodological approaches discussed earlier in this chapter are not bias free. Word embeddings have been shown to carry societal biases that are encoded in human language (Garg et al., 2018). These range from biased analogies (Bolukbasi et al., 2016; Manzini et al., 2019; Nissim et al., 2019) to bias in language ID (Blodgett and O'Connor, 2017), natural-language inference (Rudinger et al., 2017), coreference resolution (Rudinger et al., 2018), and automated essay scoring (Amorim et al., 2018).

There are corresponding efforts to reduce algorithmic bias in deep neural-network applications, for example through postprocessing (Bolukbasi et al., 2016) or directly modeling the problem (Zhao et al., 2018). However, the bias still remains encoded implicitly (Gonen and Goldberg, 2019), and transparency and awareness about the problem may be better as a research and deployment strategy (Caliskan et al., 2017; Dwork et al., 2012; Gonen and Goldberg, 2019).

There are legitimate concerns about algorithmic bias and discrimination, algorithmic accountability and transparency, and general ‘black box’ perception of deep neural-network models (Knight, 2017; Mayernik, 2017). In order to address these issues, scholars (Fiesler and Proferes, 2018; Mittelstadt et al., 2016; Olhede and Wolfe, 2018; Prates et al., 2018), AI technologists, international organizations (European Group on Ethics in Science and New Technologies (EGE), 2018), and national governments (House of Lords Select Committee, 2018) have been recently advocating for a more ‘ethical’ and ‘beneficial’ AI that will be programmed to have humans’ interests at heart and could never hurt anyone.

Kusner et al. (2017), for example, provide an ethical framework for machine decision-making, whereby a ‘decision is considered fair towards an individual if it is the same in both the actual world and a “counterfactual” world, where the individual would belong to a different demographic group’. In addition, it is vital to think about who is being excluded from AI systems and what is missing from the datasets that drive machine learning algorithms. Often, these blind spots tend to produce disparate impacts on vulnerable and marginalized groups. This leads to the invisibility of these communities and their needs because there are not enough feedback loops for individuals to give their input. While the collection of even more personal data might make algorithmic models better, it would also increase the threats to privacy.

Russell et al. (2015) present relevant questions to be considered: what are the power dynamics between different industry and research groups? Will the interests of the research community change with greater state funding? Will government intervention encourage AI research to become less transparent and accountable? What organizational principles and institutional mechanisms exist to best promote beneficial AI? What would international cooperation look like in the research, regulation, and use of AI? Will transnational efforts to regulate AI fall to the same collective-action problems that have undermined global efforts to address climate change?

To ensure that future iterations of the ethical principles are adopted widely around the world, further research will be needed to investigate long-standing political questions such as collective action, power, and governance, as well as the global governance of AI, to name a few.

# References

Ajao, O., Bhowmik, D. and Zargari, S. (2018) Fake News Identification on Twitter with Hybrid CNN and RNN Models. In: Proceedings of the 9th International Conference on Social Media and Society – SMSociety, Copenhagen, Denmark, pp. 226–230. New York: ACM.

Aletras, N., Tsarapatsanis, D., Preoţiuc-Pietro, D. and Lampos, V. (2016) Predicting judicial decisions of the European Court of Human Rights: A natural language processing perspective. PeerJ Computer Science 2: e93.

Amorim, E., Cançado, M. and Veloso, A. (2018) Automated Essay Scoring in the Presence of Biased Ratings. In: Proceedings of the 2018 Conference of the North American Chapter of the Association for Computational Linguistics: Human Language Technologies, vol. 1 (Long Papers), New Orleans, United States: Association for Computational Linguistics, pp. 229–237. Available at: https://doi.org/10.18653/v1/N18-1021 (accessed 17 December 2018).

Angwin, J., Larson, J., Mattu, S. and Kirchner, L. (2016) Machine Bias. *ProPublica*, 23 May, 2016. Available at: https://www.propublica.org/article/machine-bias-risk-assessments-in-criminal-sentencing (accessed 17 December 2018).

Atanasova, P., Barron-Cedeno, A., Elsayed, T., Suwaileh, R., Zaghouani, W., Kyuchukov, S., Da San Martino, G. and Nakov, P. (2018) Overview of the CLEF-2018 CheckThat! Lab on Automatic Identification and Verification of Political Claims. Task 1: Check-Worthiness. arXiv:1808.05542 [cs]. Available at: http://arxiv.org/abs/1808.05542 (accessed 19 December 2018).

Ba, J., Mnih, V. and Kavukcuoglu, K. (2014) Multiple Object Recognition with Visual Attention. arXiv:1412.7755 [cs]. Available at: http://arxiv.org/abs/1412.7755 (accessed 19 December 2018).

Babenko, B., Hersh, J., Newhouse, D., Ramakrishnan, A. and Swartz, T. (2017) Poverty Mapping Using Convolutional Neural Networks Trained on High and Medium Resolution Satellite Images, With an Application in Mexico. arXiv:1711.06323 [cs, stat]. Available at: http://arxiv.org/abs/1711.06323 (accessed 18 December 2018).

Baker, A. (2015) Race, paternalism, and foreign aid: Evidence from US public opinion. American Political Science Review 109(1): 93–109.

BakIr, G. (ed.) (2007) Predicting Structured Data. Advances in Neural Information Processing Systems. Cambridge, MA: MIT Press.

Barberá, P. (2015) Birds of the same feather tweet together: Bayesian ideal point estimation using Twitter data. Political Analysis 23(1): 76–91.

Baturo, A., Dasandi, N. and Mikhaylov, S. J. (2017) Understanding state preferences with text as data: Introducing the un general debate corpus. Research & Politics 4(2): 1-9. Available at https://journals.sagepub.com/doi/pdf/10.1177/2053168017712821 (accessed 19 December 2019)

Bengio, Y., Simard, P. and Frasconi, P. (1994) Learning long-term dependencies with gradient descent is difficult. IEEE Transactions on Neural Networks 5(2): 157–166.

Bengio, Y., Schwenk, H., Senécal, J.-S., Morin, F. and Gauvin, J.-L. (2006) Neural Probabilistic Language Models. In: Holmes, D. E. and Jain, L. C. (eds), Innovations in machine learning. Berlin/Heidelberg: Springer-Verlag, pp. 137–186.

Benjamins, V. R., Selic, B, Casanovas, P., Breuker, J. and Gangemi, A. (2005) Law and the Semantic Web: Legal Ontologies, Methodologies, Legal Information Retrieval, and Applications. Berlin Heidelberg: Springer.

Benoit, K. and Nulty, P. (2013) Classification methods for scaling latent political traits. In: Presentation at the Annual Meeting of the Midwest Political Science Association, Chicago, United States, pp. 11–13.

Bilbao-Jayo, A. and Almeida, A. (2018) Automatic political discourse analysis with multi-scale convolutional neural networks and contextual data. International Journal of Distributed Sensor Networks 14(11): 1-11.

Bishop, C. M. (2006) Pattern Recognition and Machine Learning. Information Science and Statistics. New York: Springer.

Blodgett, S. L. & O'Connor, B. (2017) Racial Disparity in Natural Language Processing: A Case Study of Social Media African-American English. In: 2017 Workshop on Fairness, Accountability, and Transparency in Machine Learning, Nova Scotia, Canada. arXiv preprint arXiv:1707.00061: 1-4.

Blommaert, J. and Bulcaen, C. (2000) Critical discourse analysis. Annual Review of Anthropology 29, : 447–466.

Bojanowski, P., Grave, E., Joulin, A. and Mikolov, T. (2016) Enriching Word Vectors with Subword Information. arXiv:1607.04606 [cs]. Available at: http://arxiv.org/abs/1607.04606 (accessed 19 December 2018).

Bolukbasi, T., Chang, K.-W., Zou, J. Y., Saligrama, V. and Kalai, A. T. 2016. Man is to Computer Programmer as Woman is to Homemaker? Debiasing Word Embeddings. In: Lee, D. D., von Luxburg, R. Garnett, M. Sugiyama, and Guyon, I, (eds). *Advances in Neural Information Processing Systems 29: 30th Annual Conference on Neural Information Processing Systems 2016: Barcelona, Spain, 5-10 December 2016*, Red Hook, NY: Curran Associates, Inc.: 4349–4357.

Bonica, A. (2014) Mapping the ideological marketplace. American Journal of Political Science 58(2): 367–386.

Brynjolfsson, E., Mitchell, T. and Rock, D. (2018) What Can Machines Learn, and What Does It Mean for Occupations and the Economy? In: AEA Papers and Proceedings, Nashville, TN: American Economic Association, 108: 43–47.

Bullock, J. and Luengo-Oroz, M. (2019) Automated Speech Generation from UN General Assembly Statements: Mapping Risks in AI Generated Texts. In: The 2019 International Conference on Machine Learning AI for Social Good Workshop, Long Beach, United States: 1-5. Available at http://arxiv.org/abs/1906.01946v1 (accessed 15 October 2019).

Caliskan, A., Bryson, J. J. and Narayanan, A. (2017) Semantics derived automatically from language corpora contain human-like biases. Science 356(6334): 183–186.

Ceron, A., Curini, L., Iacus, S. M. and Porro, G. (2014) Every tweet counts? How sentiment analysis of social media can improve our knowledge of citizens' political preferences with an application to Italy and France. New Media & Society 16(2): 340–358.

Danks, D. and London, A. J. (2017) Algorithmic bias in autonomous systems. In: Proceedings of the Twenty-Sixth International Joint Conference on Artificial Intelligence (IJCAI-17), Melbourne Australia. Red Hook, NY: Curran Associates, Inc: 4691–4697.

de Vries, E., Schoonvelde, M. and Schumacher, G. (2018) No longer lost in translation: Evidence that Google Translate works for comparative bag-of-words text applications. Political Analysis 26(4): 417–430.

Deng, L. (2014) Deep learning: Methods and applications. Foundations and Trends® in Signal Processing 7(3–4): 197–387.

Devlin, J., Chang, M.-W., Lee, K. and Toutanova, K. (2019) BERT: Pre-training of Deep Bidirectional Transformers for Language Understanding. In: Proceedings of the 2019 Conference of the North Americal Chapter of the Association for Computational Linguistics: Human Language Technologies, Vol 1 (Long and Short papers), Minneapolis, Minnesota: Association for Computational Linguistics: 4171 – 4186.

Dwork, C., Hardt, M., Pitassi, T., Reingold, O. and Zemel, R. (2012) Fairness through awareness. In: Proceedings of the 3rd Innovations in Theoretical Computer Science Conference, Cambridge, Massachusetts: ACM. pp. 214–226.

European Commission (2019) A Definition of AI: Main Capabilities and Scientific Disciplines. *High-Level Expert Group on Artificial Intelligence*, 8 April. Available at https://web.archive.org/web/20191014134019/https://ec.europa.eu/digital-single-market/en/news/definition-artificial-intelligence-main-capabilities-and-scientific-disciplines (accessed 14 October 2019).

European Group on Ethics in Science and New Technologies (EGE) (2018) Statement on Artificial Intelligence, Robotics and 'Autonomous' Systems. EU report. Available at: https://web.archive.org/web/20191014135156/http://ec.europa.eu/research/ege/pdf/ege_ai_statement_2018.pdf (accessed 14 October 2019).

Evans, M., McIntosh, W., Lin, J. and Cates, C. L. (2007) Recounting the courts? Applying automated content analysis to enhance empirical legal research. Journal of Empirical Legal Studies 4(4): 1007–1039.

Fairclough, N. (1989) Language and Power. London; New York: Longman.

Fairclough, N. (1992) Discourse and text: Linguistic and intertextual analysis within discourse analysis. Discourse & Society 3(2): 193–217.

Fenech, M., Strukelj, N. and Buston, O. (2018) Ethical, Social and Political Challenges of Artificial Intelligence in Health. London: Wellcome Trust and Future Advocacy. Available at: https://wellcome.ac.uk/sites/default/files/ai-in-health-ethical-social-political-challenges.pdf (accessed 21 December 2018).

Fiesler, C. and Proferes, N. (2018) 'Participant' Perceptions of Twitter Research Ethics. Social Media + Society 4(1): 1-14.

Firth, J. (1957) A synopsis of linguistic theory 1930–1955. In: Studies in Linguistic Analysis. Oxford: Philological Society.

Garg, N., Schiebinger, L., Jurafsky, D. and Zou, J. (2018) Word embeddings quantify 100 years of gender and ethnic stereotypes. Proceedings of the National Academy of Sciences, 115(16): E3635–E3644.

Goldberg, D. E. and Holland, J. H. (1988) Genetic algorithms and machine learning. Machine Learning 3(2): 95–99.

Gonen, H. and Goldberg, Y. (2019) Lipstick on a pig: Debiasing methods cover up systematic gender biases in word embeddings but do not remove them. arXiv preprint arXiv:1903.03862.

Goodfellow, I., Bengio, Y. and Courville, A. (2016) Deep Learning. Cambridge, Massachusetts London, England: MIT Press.

Gregor, K., Danihelka, I., Graves, A., Jimenez Rezende, D. and Wierstra, D. (2015) DRAW: A Recurrent Neural Network For Image Generation. arXiv:1502.04623 [cs]. Available at: http://arxiv.org/abs/1502.04623 (accessed 19 December 2018).

Grimmer, J. and Stewart, B. M. (2013) Text as data: The promise and pitfalls of automatic content analysis methods for political texts. Political Analysis 21(3): 267–297.

Helber, P., Gram-Hansen, B., Varatharajan, I., Azam, F., Coca-Castro, A., Kopackova, V. and Bilinski, P. (2018) Mapping Informal Settlements in Developing Countries with Multi-resolution, Multi-spectral Data. arXiv:1812.00812 [cs, stat]. Available at: http://arxiv.org/abs/1812.00812 (accessed 18 December 2018).

Herzog, A. and Benoit, K. (2015) The most unkindest cuts: Speaker selection and expressed government dissent during economic crisis. The Journal of Politics 77(4): 1157–1175.

Hochreiter, S. and Schmidhuber, J. (1997) Long short-term memory. Neural Computation 9(8): 1735–80.

Holmes, D. E. and Jain, L. C. (eds) (2006) Innovations in Machine Learning: Theory and Applications. Studies in Fuzziness and Soft Computing 194. Berlin: Springer.

House of Lords Select Committee (2018) AI in the UK: ready, willing and able? *House of Lords 36*. Available at: https://publications.parliament.uk/pa/ld201719/ldselect/ldai/100/100.pdf (accessed 14 October 2019).

Howard, J. and Ruder, S. (2018) Universal Language Model Fine-tuning for Text Classification. In: Proceedings of the 56th Annual Meeting of the Association for Computational Linguistics (Vol. 1: Long Papers), Melbourne, Australia: ACL, pp. 328–339.

Iyyer, M., Enns, P., Boyd-Graber, J. L. and Resnik, P. (2014) Political Ideology Detection Using Recursive Neural Networks. In: Proceedings of the 52nd Annual Meeting of the Association for Computational Linguistics, Baltimore, United States: ACL, pp. 1113–1122.

Jacovi, A., Shalom, O. S. and Goldberg, Y. (2018) Understanding Convolutional Neural Networks for Text Classification. arXiv:1809.08037 [cs]. Available at: http://arxiv.org/abs/1809.08037 (accessed 19 December 2018).

Jean, N., Burke, M., Xie, M., Davis, W. M., Lobell, D. B. and Ermon, S. (2016) Combining satellite imagery and machine learning to predict poverty. Science 353(6301): 790–794.

Johnson, R. and Zhang, T. (2014) Effective Use of Word Order for Text Categorization with Convolutional Neural Networks. arXiv:1412.1058 [cs, stat]. Available at: http://arxiv.org/abs/1412.1058 (accessed 25 October 2018).

Jordan, M. I. (2019) Artificial intelligence – The revolution hasn't happened yet. Harvard Data Science Review (1). Available at https://doi.org/10.1162/99608f92.f06c6e61 (accessed 14 October 2019).

Kalchbrenner, N., Grefenstette, E. and Blunsom, P. (2014) A Convolutional Neural Network for Modelling Sentences. arXiv:1404.2188 [cs]. Available at: http://arxiv.org/abs/1404.2188 (accessed 25 October 2018).

Kim, Y. (2014) Convolutional Neural Networks for Sentence Classification. arXiv:1408.5882 [cs]. Available at: http://arxiv.org/abs/1408.5882 (accessed 25 October 2018).

Knight, W. (2017, April) The dark secret at the heart of AI: No one really knows how the most advanced algorithms do what they do-that could be a problem. MIT Technology Review 120(2). Available at: https://www.technologyreview.com/s/604087/the-dark-secret-at-the-heart-of-ai/ (accessed 14 October 2019).

Krizhevsky, A., Sutskever, I. and. Hinton, G. E. (2012) ImageNet Classification with Deep Convolutional Neural Networks. In: Pereira, F., Burges, C. J. C., Bottou, L., and Weinberger, K.Q. (eds). *Advances in Neural Information Processing Systems 25: Neural Information Processing Systems 2012,* Red Hook, NY: Curran Associates, Inc, pp. 1097–1105.

Kudugunta, S. and Ferrara, E. (2018) Deep neural networks for bot detection. Information Sciences 467: 312–322.

Kusner, M. J., Loftus, J. R., Russell, C. and Silva, R. (2017) Counterfactual Fairness. arXiv:1703.06856 [cs, stat]. Available at: http://arxiv.org/abs/1703.06856 (accessed 18 December 2018).

Laclau, E. and Mouffe, C. (1985) Hegemony and Socialist Strategy: Towards a Radical Democratic Politics, 1st ed.: Radical Thinkers. London; New York: Verso.

Lafferty, J. D., McCallum, A. and Pereira, F. C. N (2001) Conditional Random Fields: Probabilistic Models for Segmenting and Labeling Sequence Data. In: Proceedings of the Eighteenth International Conference on Machine Learning, San Francisco, United States: Morgan Kaufmann Publishers Inc., pp. 282–289.

Lauderdale, B. E. and Herzog, A. (2016) Measuring political positions from legislative speech. Political Analysis 24(3): 374–394.

Laver, M. (2014) Measuring policy positions in political space. Annual Review of Political Science 17(1): 207–223.

Laver, M., Benoit, K. and Garry, J. (2003) Extracting policy positions from political texts using words as data. American Political Science Review 97(2): 311–331.

Lehmann, P., Werner, K., Lewandowski, J., Matthieß, T., Merz, N., Regel, S. and Werner, A. (2018) Manifesto Corpus. Version: 2018-01. Berlin: WZB Berlin Social Science Center. Available at: https://manifesto-project.wzb.eu (accessed 14 October 2019)

Liu, N. F., Gardner, M. Belinkov, Y., Peters, M. and Smith, N. A. (2019) Linguistic Knowledge and Transferability of Contextual Representations. In: NAACL 2019, Minneapolis, United

States. arXiv preprint arXiv:1903.08855. Available at https://arxiv.org/abs/1903.08855 (accessed 14 October 2019).

Makino, K., Takei, Y., Miyazaki, T. and Goto, J. (2018) Classification of Tweets about Reported Events using Neural Networks. In: Proceedings of the 2018 EMNLP Workshop W-NUT: The 4th Workshop on Noisy User-generated Text, Brussels, Belgium: Association for Computational Linguistics, pp. 153–163. Available at: http://aclweb.org/anthology/W18-6121 (accessed 14 October 2019).

Manzini, T., Lim, Y. C., Tsvetkov, Y. and Black, A. W. (2019) Black is to criminal as Caucasian is to police: Detecting and removing multiclass bias in word embeddings. In: NAACL 2019, Minneapolis, United States, pp. 1-5. arXiv preprint arXiv:1904.04047. Available at https://arxiv.org/abs/1904.04047 (accessed 14 October 2019).

Mayernik, M. S. (2017) Open data: Accountability and transparency. Big Data & Society 4(2), pp. 1-5. Available at https://journals.sagepub.com/doi/pdf/10.1177/2053951717718853 (accessed 14 October 2019).

Mikolov, T., Sutskever, I., Chen, K., Corrado, G. and Dean, J. (2013) Distributed Representations of Words and Phrases and their Compositionality. arXiv:1310.4546 [cs, stat]. Available at: http://arxiv.org/abs/1310.4546 (accessed 19 December 2018).

Mitchell, T. M. (1997) *Machine Learning*. New York: McGraw-Hill.

Mittelstadt, B. D., Allo, P., Taddeo, M., Wachter, S. and Floridi, L. (2016) The ethics of algorithms: Mapping the debate. Big Data & Society 3(2), pp. 1-21. Available at https://doi.org/10.1177/2053951716679679 (accessed 14 October 2019).

Nallapati, R., Zhou, B., dos Santos, C. N., Gulcehre, C. and Xiang, B. (2016) Abstractive Text Summarization Using Sequence-to-Sequence RNNs and Beyond. arXiv:1602.06023 [cs]. Available at: http://arxiv.org/abs/1602.06023 (accessed 25 October 2018).

Narayan, S., Cohen, S. B. and Lapata, M. (2018) Don't Give Me the Details, Just the Summary! Topic-Aware Convolutional Neural Networks for Extreme Summarization. arXiv:1808.08745 [cs]. Available at: http://arxiv.org/abs/1808.08745 (accessed 19 December 2018).

Nguyen, V.-A., Boyd-Graber, J., Resnik, P. and Miler, K. (2015) Tea Party in the House: A Hierarchical Ideal Point Topic Model and its Application to Republican Legislators in the 112th Congress. In: Proceedings of the 53rd Annual Meeting of the Association for Computational Linguistics and the 7th International Joint Conference on Natural Language Processing (vol. 1: Long Papers), Beijing, China: Association for Computational Linguistics, pp. 1438–1448.

Nissim, M., van Noord, R. and van der Goot, R. (2019) Fair is Better than Sensational: Man is to Doctor as Woman is to Doctor. arXiv preprint arXiv:1905.09866. Available at https://arxiv.org/abs/1905.09866 (accessed 14 October 2019).

Olah, C. (2015) Understanding LSTM Networks. Available at: http://colah.github.io/posts/2015-08-Understanding-LSTMs/ (accessed 21 December 2018).

Ohah, C., Mordvintsev, A. and Schubert, L. (2017) Feature Visualization. *Distill*. Available at https://distill.pub/2017/feature-visualization (accessed 14 October 2019).

Olhede, S. C. and Wolfe, P. J. (2018) The Growing Ubiquity of Algorithms in Society: Implications, Impacts and Innovations. In: Philosophical Transactions of the Royal Society,

Series A: Mathematical, Physical, and Engineering Sciences 376(2128), pp. 1 – 16. Available at https://royalsocietypublishing.org/doi/pdf/10.1098/rsta.2017.0364 (accessed 14 October 2019).

Pennington, J., Socher, R. and Manning, C. D. (2014) Glove: Global Vectors for Word Representation. In: EMNLP, Doha, Qatar.

Peters, M. E., Neumann, M., Iyyer, M., Gardner, M., Clark, C., Lee, K. and Zettlemoyer, L. (2018). Deep Contextualized Word Representations. In: NAACL, New Orleans, United States.

Peterson, A. and Spirling, A. (2018) Classification accuracy as a substantive quantity of interest: Measuring polarization in Westminster systems. Political Analysis 26(1): 120–128.

Prates, M., Avelar, P. and Lamb, L. C. (2018) On Quantifying and Understanding the Role of Ethics in AI Research: A Historical Account of Flagship Conferences and Journals. arXiv:1809.08328 [cs]: 188–173. DOI: 10.29007/74gj.

Preoţiuc-Pietro, D., Liu, Y., Hopkins, D. and Ungar, L. (2017) Beyond Binary Labels: Political Ideology Prediction of Twitter Users. In: Proceedings of the 55th Annual Meeting of the Association for Computational Linguistics (vol. 1: Long Papers), Vancouver, Canada, pp. 729–740.

Proksch, S.-O. and Slapin, J. B. (2009) How to avoid pitfalls in statistical analysis of political texts: The case of Germany. German Politics 18(3): 323–344.

Proksch, S.-O., Lowe, W. and Soroka, S. (2015) Multilingual sentiment analysis: A new approach to measuring conflict in parliamentary speeches. Legislative Studies Quarterly 44(1): 97–131.

Radford, A., Narasimhan, K., Salimans, T. and Sutskever, I. (2018) Improving language understanding by generative pre-training. OpenAI.

Radford, A., Wu, J., Child, R., Luan, D., Amodei, D. and Sutskever, I. (2019) Language models are unsupervised multitask learners. OpenAI Blog 1(8).

Ramisa, A., Yan, F., Moreno-Noguer, F. and Mikolajczyk, K. (2018) BreakingNews: Article annotation by image and text processing. IEEE Transactions on Pattern Analysis and Machine Intelligence 40(5): 1072–1085.

Rao, A. and Spasojevic, N. (2016) Actionable and Political Text Classification using Word Embeddings and LSTM. Available at: https://arxiv.org/abs/1607.02501 (accessed 29 November 2018).

Ruder, S., Peters, M., Swayamdipta, S. and Wolf T. (2019) Transfer Learning in Natural Language Processing. In: Proceedings of the 2019 Conference of the North American Chapter of the Association for Computational Linguistics, pp. 15–18.

Rudinger, R., May, C. and Van Durme, B. (2017) Social Bias in Elicited Natural Language Inferences. In: Proceedings of the First ACL Workshop on Ethics in Natural Language Processing, pp. 74–79.

Rudinger, R., Naradowsky, J., Leonard, B. and Van Durme, B. (2018) Gender Bias in Coreference Resolution. arXiv preprint arXiv:1804.09301.

Russell, S., Dewey, D. and Tegmark, M. (2015) Research priorities for robust and beneficial artificial intelligence. AI Magazine 36(4): 105–114.

Samothrakis, S. (2018) Viewpoint: Artificial Intelligence and Labour. arXiv:1803.06563 [cs]. Available at: https://www.ijcai.org/proceedings/2018/0803.pdf (accessed 12 December 2018).

Samuel, A. L. (1959) Some studies in machine learning using the game of checkers. IBM Journal of Research and Development 3(3): 210–229.

Sanders, J., Lisi, G. and Schonhardt-Bailey, C. (2017) Themes and topics in parliamentary oversight hearings: a new direction in textual data analysis. Statistics, Politics and Policy 8(2): 153–194.

Schlogl, L. and Sumner, A. (2018) The Rise of the Robot Reserve Army: Automation and the Future of Economic Development, Work, and Wages in Developing Countries. ID 3208816, SSRN Scholarly Paper, 2 July. Rochester, NY: Social Science Research Network. Available at: https://papers.ssrn.com/abstract=3208816 (accessed 19 December 2018).

Singh, V., Varshney, A., Akhtar, S. S. Vijay, D. and Shrivastava, M. (2018) Aggression Detection on Social Media Text Using Deep Neural Networks. In: Proceedings of the 2nd Workshop on Abusive Language Online (ALW2), Brussels, Belgium, pp. 43–50. Association for Computational Linguistics. Available at: http://aclweb.org/anthology/W18-5106.

Slapin, J. B. and Proksch, S.-O. (2008) A scaling model for estimating time-series party positions from texts. American Journal of Political Science 52(3): 705–722.

Spirling, A. and Rodriguez, P. L. (2019) Word Embeddings: What Works, What Doesn't, and How to Tell the Difference for Applied Research. NYU manuscript. https://github.com/ArthurSpirling/EmbeddingsPaper

Tenney, I., Das, D. and Pavlick, E. (2019) BERT Rediscovers the Classical NLP Pipeline. ACL 2019.

Uysal, A. K. (2016) An improved global feature selection scheme for text classification. Expert Systems with Applications 43: 82–92.

Wang, Y. and Kosinski, M. (2018) Deep neural networks are more accurate than humans at detecting sexual orientation from facial images. Journal of Personality and Social Psychology, 114(2): 246.

Yamshchikov, I. P. and Rezagholi, S. (2018) Elephants, Donkeys, and Colonel Blotto. In: Proceedings of the 3rd International Conference on Complexity, Future Information Systems and Risk, pp. 113–119.

Yang, Z., Dai, Z., Yang, Y., Carbonell, J., Salakhutdinov, R. and Le, Q. V. (2019) XLNet: Generalized Autoregressive Pretraining for Language Understanding. arXiv 1906.08237.

Yu, B., Kaufmann, S. and Diermeier, D. (2008) Classifying party affiliation from political speech. Journal of Information Technology & Politics 5(1): 33–48.

Zhang, X., Zhao, J. and LeCun, Y. (2015) Character-level Convolutional Networks for Text Classification. arXiv:1509.01626 [cs]. Available at: http://arxiv.org/abs/1509.01626 (accessed 25 October 2018).

Zhang, Y. and Wallace, B. (2015) A Sensitivity Analysis of (and Practitioners' Guide to) Convolutional Neural Networks for Sentence Classification. Available at: https://arxiv.org/abs/1510.03820 (accessed 25 October 2018).

Zhang, Q. and Zhu, S. (2018) Visual interpretability for deep learning: A survey. Frontiers of Information Technology and Electronic Engineering 19(1): 27–39.

Zhang, Q., Wang, W. and Zhu, S. C. (2018) Examining CNN Representations with Respect to Dataset Bias. In: 32nd AAAI Conference on Artificial Intelligence.

Zhao, J., Zhou, Y. Li, Z., Wang, W. and Chang, K.-W. 2018. Learning Gender-Neutral Word Embeddings. In: Proceedings of the 2018 Conference on Empirical Methods in Natural Language Processing, Brussels, Belgium, pp. 4847–4853, Association for Computational Linguistics.